\documentclass{article}

    \PassOptionsToPackage{numbers, compress}{natbib}

 \usepackage[main, final]{neurips_2026}

\usepackage[utf8]{inputenc} 
\usepackage[T1]{fontenc}    
\usepackage{hyperref}       
\usepackage{url}            
\usepackage{booktabs}       
\usepackage{amsfonts}       
\usepackage{nicefrac}       
\usepackage{microtype}      
\usepackage{xcolor}         
\usepackage{multirow}
\usepackage{adjustbox}
\usepackage{makecell}
\usepackage{caption}
\usepackage{rotating}   
\usepackage{amsmath} 
\usepackage{enumitem} 
\usepackage{graphicx}
\usepackage{wrapfig}

\usepackage[dvipsnames]{xcolor}
\newcommand{\gain}[1]{{\scriptsize\textcolor{BrickRed}{(+#1)}}}
\newcommand{\std}[1]{{\scriptsize\textcolor{ForestGreen}{$\pm$#1}}}
\newcommand{\ostd}[1]{{\scriptsize\textcolor{gray}{$\pm$#1}}}

\definecolor{oracle}{gray}{0.45}
\newcommand{\oraclerow}[1]{\textcolor{oracle}{\textit{#1}}}

\title{Preference-Guided Adaptation for Open-Vocabulary Semantic Segmentation via Prompt Disagreement}

\author{%
    Hyun-Kurl Jang\\
  Visual Intelligence Lab.\\
  KAIST\\
  \texttt{jhg0001@kaist.ac.kr} \\
  \And
  Jihun Kim\\
  Visual Intelligence Lab.\\
  KAIST\\
  \texttt{jihun1998@kaist.ac.kr} \\
  \And
  Kuk-Jin Yoon\\
  Visual Intelligence Lab.\\
  KAIST\\
  \texttt{kjyoon@kaist.ac.kr} \\
}

\begin{document}

\maketitle

\begin{abstract}
Open-vocabulary semantic segmentation (OVSS) enables pixel-level prediction over arbitrary text-specified vocabularies and has shown strong generalization on common benchmarks. However, OVSS performance often degrades in specialized domains such as medical imaging, remote sensing, and industrial inspection, where dense pixel-level masks for adaptation are costly to obtain and require domain-specific expertise. We propose a preference-guided adaptation framework that replaces dense mask supervision with binary preferences. We observe that different prompt templates produce systematically different segmentations for the same image, a phenomenon we call prompt disagreement, and we repurpose it as a built-in source of preference supervision. Building on this, we mine localized preference queries from regions of high cross-template uncertainty, and adapt the OVSS model with Region-Localized Preference Optimization (RLPO) together with consistency regularization that stabilizes updates outside the queried region. Across extensive experiments on the MESS benchmark, the proposed method achieves consistent gains across diverse OVSS backbones without any pixel-level annotation, and remains effective under noisy preferences. Our code is available at \url{https://github.com/blue-531/pref-ovss}.
\end{abstract}

\section{Introduction}
\label{sec:intro}
\vspace{-5pt}
Semantic segmentation has long been studied under a closed-set assumption, where models are trained and evaluated on a fixed set of categories~\cite{deeplabv2,deeplabv3,deepabv3plus,pspnet,hrnet,maskformer,cheng2022masked,segformer,ocrnet,kim2026bootstrapping}.
This limits their ability to recognize novel concepts and operate in open-world settings.
Open-vocabulary semantic segmentation (OVSS)~\cite{li2022language,ghiasi2022scaling, xu2022simple, luo2023segclip, ding2022open, liang2023open, zou2023generalized, xu2023side, xu2023open, xu2023masqclip, chen2023open, wang2023hierarchical, jiao2023learning, ma2023attrseg, yu2023convolutions, qin2023freeseg, han2023open, zhang2023simple, xu2024transferable, cho2024cat, xie2024sed, shan2024open, jiao2024collaborative, zhou2023zegclip, li2025mask, niu2025eov, qorbani2025semantic, zhao2025dpseg, yoon2026dinode} addresses this limitation by allowing class names to be specified in natural language at inference time.
Powered by vision-language models (VLMs) such as CLIP~\cite{radford2021learning}, OVSS aligns dense visual features with text embeddings and has shown strong generalization to categories unseen during training.

As OVSS moves toward real-world deployment, adapting it to specialized domains has become increasingly important.
Applications such as medical imaging, remote sensing, industrial inspection, and agriculture differ substantially from web-scale pretraining data in appearance, label granularity, and vocabulary~\cite{zou2023generalized,MESSBenchmark2023}.
Prior adaptation strategies for OVSS have explored prompt tuning~\cite{yilmaz2024opendas,adhikari2024tunevlseg,poudel2023exploring} and adapter-based fine-tuning~\cite{dhakal2024vlsm,mishra2026improvise}. However, many of these approaches often assume access to target-domain mask supervision, which are difficult to obtain in specialized domains.
Unlike common-object datasets, specialized domains often involve fine-grained categories, ambiguous visual boundaries, and domain-specific terminology, making reliable mask annotation dependent on expert knowledge.
Since target classes and annotation criteria often vary across specialized domains, relying on dense masks creates a recurring annotation bottleneck for OVSS adaptation.


These limitations motivate a different form of supervision that does not 
require annotators to construct dense ground truth yet still guides model 
predictions toward the intended concept.
Pairwise preferences provide such a signal.
Preference-based supervision has emerged as an effective way to align model outputs with human intent in language and vision-language generation, largely because relative judgments can convey useful training signals without requiring fully specified target outputs~\cite{christiano2017deep,rafailov2023direct,wallace2024diffusion}.
Translating this principle to OVSS leads to a lightweight protocol: given two candidate segmentations for the same image, annotators simply choose the prediction that better matches the intention.
By indicating which prediction better matches the intended segmentation in a target-domain image, this relative feedback provides a natural form of supervision for adapting OVSS models without requiring pixel-level masks.
\begin{figure}[t]
\centering
\includegraphics[width=0.99\linewidth]{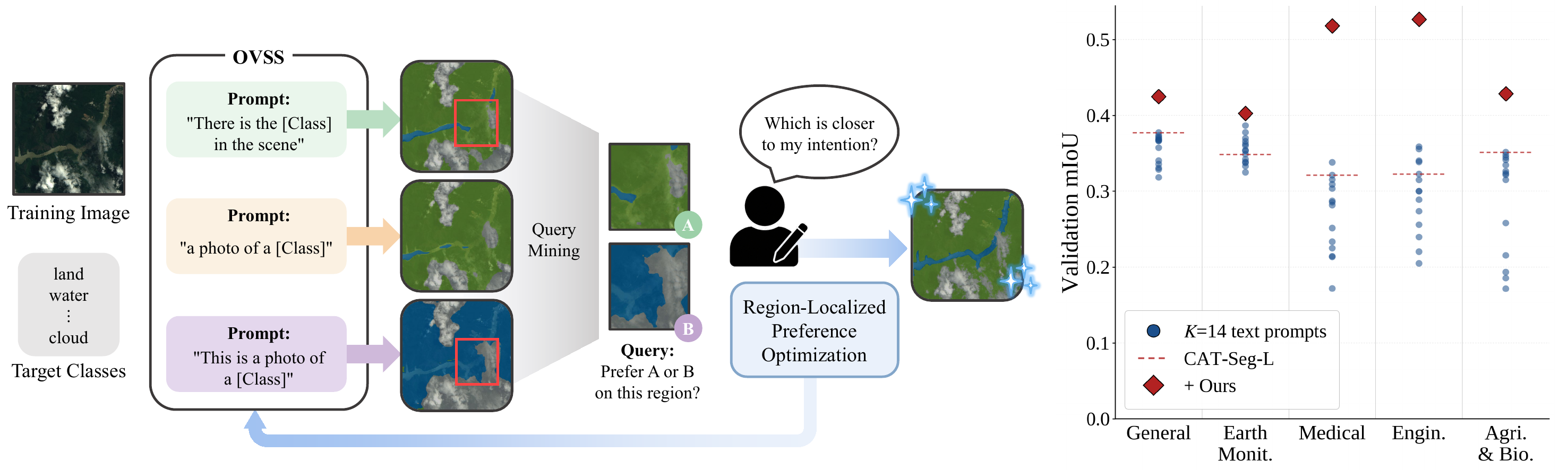}
\caption{\textbf{Left:} For each target-domain image, our framework runs multiple prompt templates through a OVSS backbone, mines a localized preference query from regions where the templates disagree most, and uses a binary judgment ("which prediction is closer to the intended concept on this region?") to drive an update of OVSS model. \textbf{Right:} Validation mIoU across diverse specialized domains~\cite{MESSBenchmark2023} when only the prompt template is varied. Across all domains, performance varies widely across templates. Our method harnesses this variance as supervision and lifts performance above the entire per-domain spread, turning prompt disagreement into training signal.}
\label{fig:motiv}
\vspace{-20pt}
\end{figure}

To make this supervision actionable, however, OVSS requires meaningful candidate segmentations to compare.
We observe that such candidates naturally arise from the prompting interface itself.
Different prompt templates can produce systematically different masks for the same image and class name~\cite{radford2021learning,gu2021open,zhou2022learning}; we refer to this template-induced prediction variance as \textbf{prompt disagreement}.
As shown in Fig.~\ref{fig:motiv} (right), prompt disagreement is pronounced in practice: across diverse specialized domains, validation mIoU varies widely across templates for the same backbone. Rather than treating prompt disagreement as a nuisance, we repurpose it as a built-in source of pairwise supervision: prompt-induced masks serve as competing hypotheses, and binary preferences identify which hypothesis better captures the intended concept.

Building on this idea, we propose a preference-guided adaptation framework for OVSS that turns prompt disagreement into an actionable training signal, illustrated in Fig.~\ref{fig:motiv} (left).
The framework consists of three components.
First, we mine informative preference queries from prompt ensembles.
We localize high-uncertainty regions using cross-prompt entropy and, within each region, select the pair of template-induced predictions with the largest disagreement.
This jointly determines where feedback should be collected and which pair of segmentation hypotheses should be compared.
Second, we introduce Region-Localized Preference Optimization (RLPO) for adapting OVSS models from binary preferences.
Rather than treating the preference as a single image-level signal, our objective applies it to pixel-level segmentation scores inside the selected region, enabling dense spatial supervision from a single comparison.
Third, we introduce a consistency regularization to stabilize preference optimization.
Because the preference objective only supervises the selected region, it resolves the queried disagreement but leaves predictions outside that region unconstrained.
Our consistency regularizer therefore uses the preferred prediction as a pseudo-target for the rejected prediction outside the selected region, preventing unintended drift beyond the adapted area.



We evaluate the proposed method on the MESS benchmark~\cite{MESSBenchmark2023}, which covers five domain groups: general scenes, earth monitoring, medical sciences, engineering, and agriculture \& biology.
Across these diverse settings, our approach consistently improves OVSS performance without dense masks, demonstrating that prompt disagreement provides a practical supervision signal for adapting OVSS to specialized domains.
We further show that the proposed adaptation strategy yields consistent improvements across different OVSS baselines~\cite{xu2023side,cho2024cat}, suggesting that it is not specific to a single model design.
The method also remains effective under noisy preference feedback, highlighting the robustness of binary preferences.

Our main contributions are as follows:
\vspace{-5pt}
\begin{itemize}
[itemsep=-1pt, leftmargin=*]
    \item We show that prompt disagreement can be repurposed as a source of preference supervision, enabling OVSS adaptation without human-provided 
pixel-level masks.

    \item We propose Region-Localized Preference Optimization (RLPO), which adapts OVSS models from binary preferences mined via prompt disagreement.

    \item We demonstrate consistent improvements on the MESS benchmark across diverse specialized domains, validating binary preference feedback as a practical supervision signal for OVSS adaptation.
\end{itemize}
\newcommand{\eg}{e.g.} 

\section{Related Works}
\label{sec:related_works}
\subsection{Open-Vocabulary Semantic Segmentation}
\label{sec:related_ovs}
 
Open-vocabulary semantic segmentation (OVSS) aims to assign pixel-level labels from an arbitrary set of text-specified classes, including those unseen during training.
Building on vision-language foundations such as CLIP~\cite{radford2021learning}, recent methods have explored diverse strategies for bridging image-level pretraining with dense prediction.
Two-stage approaches~\cite{ghiasi2022scaling,xu2022simple,ding2022open, liang2023open, xu2023open, chen2023open, jiao2023learning} first generate mask proposals and then classify each region via CLIP, while one-stage methods~\cite{xu2023side, yu2023convolutions, zhou2023zegclip} produce masks within a side adapter during inference.
Cost aggregation approaches~\cite{cho2024cat,xie2024sed, zhao2025dpseg, gandhamal2025ov} instead refine patch-level embeddings through cost aggregation or learned decoders to directly produce per-pixel predictions.

Despite strong results on standard benchmarks dominated by everyday imagery, such as ADE20K~\cite{zhou2017scene} and Pascal Context~\cite{mottaghi_cvpr14}, OVSS models often struggle in specialized domains.
Such domains often exhibit domain-specific vocabulary, high inter-class similarity, ambiguous boundaries, and atypical visual appearance.
Existing adaptation strategies, including prompt tuning~\cite{yilmaz2024opendas,adhikari2024tunevlseg,poudel2023exploring}, adapter-based methods~\cite{qorbani2025semantic,dhakal2024vlsm,mishra2026improvise}, and personalized OVSS approaches~\cite{park2025personalized}, typically rely on dense mask supervision. This reliance limits scalability in specialized domains, where annotation often requires expert knowledge and must be repeated for each target vocabulary. To address this gap, we propose a preference-guided OVSS adaptation framework that learns from binary comparisons between candidate segmentations rather than dense masks.



\subsection{Preference Learning} 
\label{sec:related_preference}

Preference learning~\cite{christiano2017deep,rafailov2023direct, jain2013learning, busa2014preference, kupcsik2017learning, sadigh2017active, azar2024general, zhang2024negative, fan2024simplicity} uses relative judgments between candidate outputs as supervision, instead of requiring fully specified ground-truth targets. This formulation is especially useful in settings where dense annotation is difficult to obtain but comparative feedback is easier to collect. Among recent approaches, Direct Preference Optimization (DPO)~\cite{rafailov2023direct} has emerged as a simple objective that learns directly from pairwise preferences without requiring an explicit reward model. 

While DPO was originally proposed for language model alignment, preference-based objectives have since been extended to visual tasks, including image generation~\cite{liang2025aesthetic, wallace2024diffusion, yang2024using, yang2024dense} and dense prediction~\cite{cai2025dspo,wu2025dp}. Recent work has also begun to explore preference-based objectives for segmentation~\cite{konwer2025enhancing, wu2025sampo}. However, these studies have been limited to fixed-target settings, typically in medical imaging, where the task involves a single foreground structure or a small closed label space. 

As a result, preference-guided adaptation remains largely unexplored for open-vocabulary semantic segmentation. We address this gap by using prompt-induced variation to construct localized binary comparisons, enabling OVSS adaptation from preference feedback without dense masks.

\section{Method}
\label{sec:method}

\begin{figure}[t]
\centering
\includegraphics[width=0.99\linewidth]{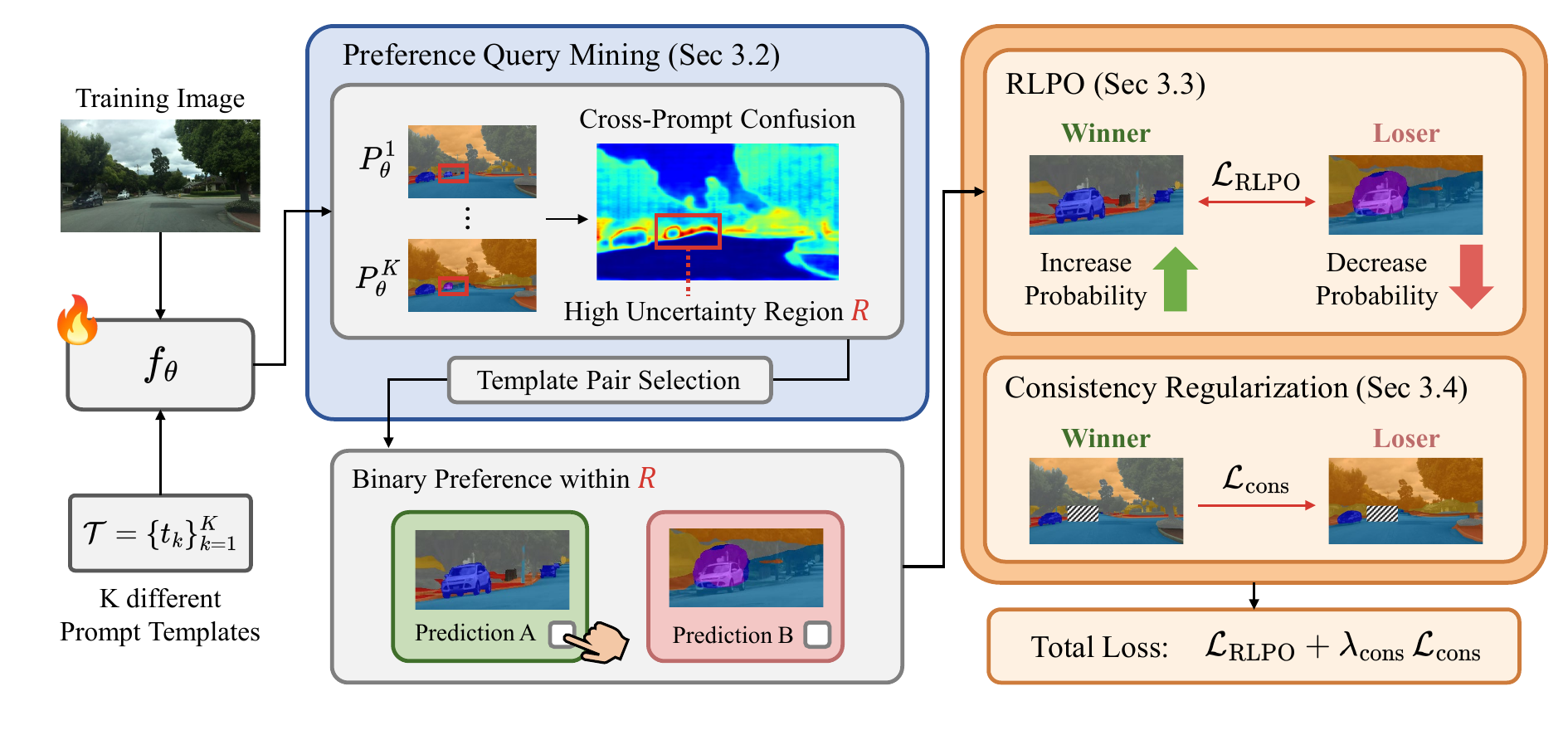}
\vspace{-10pt}
\caption{Given a training image, our method generates \(K\) prompt-conditioned predictions with model \(f_\theta\). It mines a preference query by selecting a high-uncertainty region \(R\) from prompt-ensemble entropy and the most disagreeing template pair within \(R\). A binary preference identifies the winner and loser predictions. RLPO updates \(f_\theta\) by increasing the winner's score relative to the loser's inside R, while the consistency regularization aligns the loser prediction with the winner pseudo-label.}
\label{fig:overall}
\vspace{-5pt}
\end{figure}

Our goal is to adapt an OVSS model to a specialized target domain using binary preference feedback.
Given target-domain images and a target vocabulary, we use prompt disagreement to construct localized comparison queries and obtain binary preferences over candidate segmentations.
These preferences serve as the supervision signal for adaptation, without requiring dense annotations.  Our overall framework is illustrated in Fig.~\ref{fig:overall}, which consists of three components: preference query mining, RLPO, and consistency regularization.
We first introduce the problem setting and then describe each component.

\subsection{Preliminaries}
\label{sec:prelim}
 
\paragraph{Open-vocabulary semantic segmentation (OVSS).}
OVSS aims to assign a class label to every pixel of an image given an arbitrary class vocabulary specified at test time, including categories unseen during training. Modern OVSS systems build on vision--language models such as CLIP~\cite{radford2021learning}, predicting segmentation maps by aligning dense visual features with text embeddings of class names prompted through templates (e.g., \texttt{``a photo of a [CLASS]''}). The choice of prompt template is known to materially affect predictions~\cite{radford2021learning,gu2021open,zhou2022learning}: different templates can produce noticeably different segmentations on the same image, particularly under domain shift, where the source-trained alignment is no longer well calibrated to the target distribution.

\paragraph{Direct preference optimization (DPO).}
Standard reinforcement learning from human feedback first fits a reward model on preference data and then optimizes a policy against it; DPO~\cite{rafailov2023direct} avoids the reward modeling stage by deriving a closed-form objective that learns directly from preference pairs. Specifically, DPO adapts a policy $\pi_\theta$ from a frozen reference $\pi_\text{ref}$ using preference pairs $(y_w, y_l)$, where $y_w$ is preferred to $y_l$ for the same input. It interprets the log-probability ratio
\begin{equation}
    r_\theta(y) = \log\frac{\pi_\theta(y)}{\pi_\text{ref}(y)}
    \label{eq:dpo-reward}
\end{equation}
as an implicit reward, and optimizes the Bradley--Terry objective~\cite{bradley1952rank}
\begin{equation}
    \mathcal{L}_\text{DPO}(\theta) = -\log\sigma\!\left(\beta\bigl[r_\theta(y_w) - r_\theta(y_l)\bigr]\right),
    \label{eq:dpo-loss}
\end{equation}
which raises $r_\theta(y_w)$ for the winner while lowering $r_\theta(y_l)$ for the loser. Because the reward is defined relative to $\pi_\text{ref}$, the reference distribution acts as a KL anchor that prevents $\pi_\theta$ from drifting far from its initial behavior, with $\beta$ controlling the trade-off between matching the preferences and staying close to $\pi_\text{ref}$. We adapt this principle to OVSS in Section~\ref{sec:rlpo} by replacing the policy log-probability ratio with a region-localized segmentation score, so that Eq.~(\ref{eq:dpo-loss}) takes a form directly applicable to template-conditional segmentation.

\paragraph{Problem setting.}
\label{sec:problem-setting}
We adapt an OVSS model to a target domain in a streaming, single-step setting: training images from the target domain arrive sequentially, and for each image we elicit a binary preference and perform a single gradient update on the model parameters $\theta$ before moving on, without revisiting images or accumulating preferences for batch optimization.
Let $x \in \mathbb{R}^{H \times W \times 3}$ denote a target-domain image with pixel domain $\Omega$ and pixel positions $\mathbf{u} \in \Omega$. Let $\mathcal{C} = \{c_1, \ldots, c_N\}$ denote the target vocabulary, and let $\mathcal{T} = \{t_k\}_{k=1}^{K}$ denote a fixed set of $K$ templates with prompted vocabulary $\mathcal{C}^k = \{t_k(c) : c \in \mathcal{C}\}$. 
We adapt an OVSS model $f_\theta$ from a frozen reference $f_{\mathrm{ref}}$. Both share the same backbone; $f_\theta$ adds lightweight adapters on the vision and text branches as the only trainable parameters, while $f_{\mathrm{ref}}$ corresponds to the initial state of $f_\theta$ before adaptation.
The template-$k$ pixel-level prediction is
\begin{equation}
P_\theta^k(c \mid x, \mathbf{u}) = f_\theta(x, \mathcal{C}^k)_{\mathbf{u}, c},
\qquad c \in \mathcal{C},
\end{equation}
with $P_{\mathrm{ref}}^k$ defined analogously. 
Given a small set of target-domain images, we adapt $\theta$ using binary preferences elicited between pairs of template-specific predictions $\{P_\theta^k\}_{k=1}^{K}$.

\subsection{Preference Query Mining}
\label{sec:query}

We design the preference protocol around two requirements: each query should be cognitively simple to answer, and the resulting binary signal should still carry enough supervision to drive adaptation. Richer feedback formats (e.g., ratings, rankings, or multi-way selections) place a heavier cognitive load on annotators and are prone to inconsistent calibration across examples and annotators~\cite{casper2023open}. We therefore restrict each annotation to a binary choice.
 Even with binary feedback, whole-image preferences remain ambiguous in segmentation, because two candidates may each be better in different parts of the image. Therefore, we further localize each comparison to a small region $R \subseteq \Omega$, on which the question becomes concrete: within $R$, which of two template predictions $P^a_\theta, P^b_\theta$ better matches the intended concept?

We construct queries by exploiting prompt disagreement, the variation 
across template predictions $\{P_\theta^k\}_{k=1}^{K}$.
Template-induced predictions provide useful candidates because they vary the segmentation hypothesis while preserving the target vocabulary. This makes the resulting candidates directly comparable under the same semantic target. Concretely, prompt disagreement on a single image yields two operational signals: where the templates collectively disagree, and which pair of templates disagrees most strongly. To localize regions of high disagreement, we measure cross-prompt uncertainty at each pixel by the entropy of the ensemble distribution
\begin{equation}
\bar P(c \mid x, \mathbf{u}) 
= 
\frac{1}{K} \sum_{k=1}^{K} P_\theta^k(c \mid x, \mathbf{u}),
\qquad
\mathcal{H}(\mathbf{u}) 
= 
- \sum_{c \in \mathcal{C}} \bar P(c \mid x, \mathbf{u}) 
\log \bar P(c \mid x, \mathbf{u}).
\end{equation}
We binarize $\mathcal{H}$ at its $0.95$-quantile and take $R$ as the 
bounding box of the largest connected component of the resulting 
high-entropy mask. Within $R$, we identify the most informative 
template pair by counting pixel-level disagreement between hard 
predictions and selecting the pair with the largest count:
\begin{equation}
(a, b) = \arg\max_{1 \le k < k' \le K} 
\sum_{\mathbf{u} \in R} \mathbf{1}\!\left[\,\hat Y^k(\mathbf{u}) \neq \hat Y^{k'}(\mathbf{u})\,\right],
\quad
\hat Y^k(\mathbf{u}) = \arg\max_c P_\theta^k(c \mid x, \mathbf{u}).
\end{equation}
By targeting both the most uncertain region and the most disagreeing template pair within it, each query is visually concrete enough to be judged at a glance yet carries dense supervision for adaptation.

\subsection{Region-Localized Preference Optimization (RLPO)}
\label{sec:rlpo}

Given the query $(R, a, b)$, an oracle provides a binary preference indicating which of the two predictions $P^a_\theta$ and $P^b_\theta$ is preferred on $R$. We denote the winner and loser template indices by $w$ and $l$, respectively, with corresponding hard predictions $\hat{Y}^w$ and $\hat{Y}^l$.

\paragraph{Region-level score.}
To instantiate the DPO objective in our setting, we need a per-template scalar score that plays the role of $\log \pi_\theta(y)$ in Eq.~(\ref{eq:dpo-reward}). A natural choice is the log-likelihood that template $k$ assigns to its own hard prediction, averaged over $R$. However, a simple pixelwise average would be dominated by classes that occupy the largest area in $R$, so a single large class could obscure the contribution of smaller but semantically important ones. We therefore use a class-balanced average: pixels in $R$ are grouped by the winner's prediction $\hat{Y}^w$ to define a stable, $k$-independent partition, and per-pixel scores are first averaged within each class before averaging across classes. Let
\begin{equation}
    R_c = \{\mathbf{u} \in R : \hat{Y}^w(\mathbf{u}) = c\}, \qquad U_R = \{c \in \mathcal{C} : R_c \neq \emptyset\}.
\end{equation}
For $k \in \{w, l\}$, the score is
\begin{equation}
    S^k_\theta(R) = \frac{1}{|U_R|} \sum_{c \in U_R} \frac{1}{|R_c|} \sum_{\mathbf{u} \in R_c} \log P^k_\theta\bigl(\hat{Y}^k(\mathbf{u}) \mid x, \mathbf{u}\bigr),
    \label{eq:score}
\end{equation}
and the reference score $S^k_\text{ref}(R)$ is defined analogously by replacing $P^k_\theta$ with $P^k_\text{ref}$.

\paragraph{Region-localized preference loss.}
Treating the score difference $S^k_\theta(R) - S^k_\text{ref}(R)$ as the implicit reward $r_\theta$ of Eq.~(\ref{eq:dpo-reward}), we instantiate the Bradley--Terry objective of Eq.~(\ref{eq:dpo-loss}) for template-conditional segmentation:
\begin{equation}
    \mathcal{L}_\text{RLPO}(\theta) = -\log\sigma\!\left(\beta\bigl[\bigl(S^w_\theta(R) - S^w_\text{ref}(R)\bigr) - \bigl(S^l_\theta(R) - S^l_\text{ref}(R)\bigr)\bigr]\right).
    \label{eq:rlpo}
\end{equation}
Intuitively, minimizing $\mathcal{L}_\text{RLPO}$ produces the same winner-up/loser-down dynamic as standard DPO, but applied to region-localized segmentation scores: within $R$, $f_\theta$ becomes more confident in the winner template's prediction $\hat{Y}^w$ and less confident in the loser's prediction $\hat{Y}^l$, with $f_\text{ref}$ anchoring both shifts. 
The full derivation of $\mathcal{L}_\text{RLPO}$ from the standard DPO formulation is provided in Appendix~\ref{app:rlpo}.

\subsection{Consistency Regularization}
\label{sec:cons}

The preference loss in Eq.~\eqref{eq:rlpo} only updates the model 
within $R$, leaving the loser branch unconstrained elsewhere. 
Without an additional anchor, the strong preference signal applied on 
$R$ can shift the loser's predictions outside $R$ in unintended 
directions. To prevent this, we add a consistency regularizer that 
keeps the loser-template prediction aligned with the winner pseudo-label 
$\hat Y^w$ outside $R$, restricted to pixels where the winner is 
sufficiently confident. We use the Lov\'{a}sz--Softmax 
loss~\citep{berman2018lovasz}:
\begin{equation}
\mathcal{L}_{\mathrm{cons}}(\theta) 
= 
\mathcal{L}_{\mathrm{Lov\acute{a}sz}}\!\left(
P_\theta^l\big|_{{R}_{\mathrm{conf}}},\;
\hat Y^w\big|_{{R}_{\mathrm{conf}}}
\right),
\label{eq:cons}
\end{equation}
where $R_{\mathrm{conf}}$ denotes the pixels outside $R$ at 
which confidence exceeds a threshold 
$\tau_{\mathrm{conf}}$.

The full per-shot objective combines preference and consistency:
\begin{equation}
\mathcal{L}_{\mathrm{total}}(\theta) 
= 
\mathcal{L}_{\mathrm{RLPO}}(\theta) 
+ 
\lambda_{\mathrm{cons}}\, \mathcal{L}_{\mathrm{cons}}(\theta),
\label{eq:total}
\end{equation}
where $\lambda_{\mathrm{cons}}$ balances the two terms. 
Following the streaming protocol of Section~\ref{sec:problem-setting}, each incoming image produces exactly one minimization step of $\mathcal{L}_{\mathrm{total}}(\theta)$.

\section{Experiments}
\label{sec:exp}

\subsection{Experimental Setup}
\label{sec:setup}

\paragraph{Datasets.}
We evaluate on the MESS benchmark~\cite{MESSBenchmark2023}, a suite designed 
to stress-test open-vocabulary segmentation models on domains that 
substantially differ from web-scale image-text pretraining. MESS spans five domain groups---%
General~\cite{DatasetATLANTIS,DatasetFoodSeg103,DatasetMHPv1,DatasetBDD100K}, 
Earth Monitoring~\cite{DatasetUAVid,DatasetFloodNet,DatasetWorldFloods,DatasetiSAID}, 
Medical Sciences~\cite{DatasetPAXRay4,DatasetCHASEDB1,DatasetKvasirInstrument}, 
Engineering~\cite{DatasetZeroWastef,DatasetPST900,DatasetDeepCrack,DatasetCorrosionCS}, 
and Agriculture \& Biology~\cite{DatasetCWFID,DatasetCUB200,DatasetSUIM}%
---each containing multiple datasets with distinct visual appearance, 
label granularity, and vocabulary. We exclude four datasets from the original MESS benchmark---Dark Zurich~\cite{DatasetDarkZurich}, DRAM~\cite{DatasetDRAM}, ISPRS Potsdam~\cite{DatasetISPRSPotsdam}, and CryoNuSeg~\cite{DatasetCryoNuSeg}---as they either lack a training split or are no longer publicly accessible. Following the original 
benchmark protocol, we report mean intersection-over-union (mIoU, \%) 
per domain and the overall mean across datasets; per-dataset results 
are provided in Appendix~\ref{appendix:per-dataset}. Adaptation 
samples are drawn from each dataset's training split, while 
evaluation is performed on the official held-out split.

\paragraph{Base OVSS models.}
We apply our adaptation framework to two OVSS backbones, SAN~\cite{xu2023side} and CAT-Seg~\cite{cho2024cat}, each in two CLIP-scale variants (ViT-B/16 and ViT-L/14). All backbones are kept frozen except for lightweight adapters: a LoRA module of rank $4$ on the vision branch and a residual prompt embedding on the text branch. For adaptation, we draw $K = 14$ prompt templates from the ViLD prompt pool~\cite{gu2021open}. The full template list and an analysis of 
template diversity are provided in Appendix~\ref{appendix:template-analysis}. At evaluation time, we use each backbone's default template (the ViLD pool for SAN and \texttt{``a photo of a [CLASS] in the scene''} for CAT-Seg) to match its baseline configuration.

\paragraph{Preference oracle.}
The binary preference for each query $(R, a, b)$ is provided by an 
oracle that compares the two candidate predictions to the 
ground-truth segmentation within $R$: the winner $w \in \{a, b\}$ 
is the template whose prediction has the higher 
intersection-over-union with the ground truth restricted to $R$. 
This mimics an annotator who, given two segmentations cropped to a 
small region, selects the one closer to the intended target. We verify that this is a faithful stand-in for a human annotator in 
Appendix~\ref{appendix:oracle-validation}.


\paragraph{Baselines.}
We compare against two reference points: (i)~the \emph{baseline} OVSS model with no
adaptation, evaluated zero-shot on each target domain; and (ii)~the
\emph{Dense-mask} reference, which follows the same streaming adaptation protocol but replaces the binary preference with ground-truth mask supervision, isolating the
effect of the supervision format under a matched budget rather than serving as a
fully-supervised upper bound. Unless stated otherwise,
both our method and the Dense-mask reference adapt each backbone with $64$ target
domain images per dataset\footnote{For datasets with fewer training set images,
results are reported with $8$ (CHASE DB1) and $32$ (CWFID) images.}. The effect of
varying the number of training images is studied in Section~\ref{sec:ablation}. All
adaptation results are averaged over three seeds, each with independently sampled
adaptation images. A broader comparison, against additional baselines, is
provided in Appendix~\ref{appendix:alternatives}. Additional implementation details
are provided in Appendix~\ref{appendix:implementation}.

\begin{table}[t]
\centering
\small
\setlength{\tabcolsep}{5pt}
\renewcommand{\arraystretch}{1.15}
\caption{Quantitative evaluation on the MESS benchmark (mIoU, \%). Numbers for our 
method are mean $\pm$ standard deviation. \textcolor{oracle}{\textit{Dense-mask}} rows use ground-truth mask supervision under the same streaming protocol, providing the dense-supervision counterpart.}
\vspace{3pt}
\label{tab:mess_main}
\resizebox{\linewidth}{!}{%
\begin{tabular}{llcccccc}
\toprule
\textbf{VLM} & \textbf{Method} & General & Earth Monit. & Medical Sci. & Engineering & Agri.\ \& Biology & \textbf{Mean} \\
\midrule
\multirow{6}{*}{CLIP ViT-B/16}
 & SAN-B                         & 21.60 & 26.90 & 33.51 & 28.19& 15.99 & 25.29 \\
 & \enspace \textbf{+ Ours}        & 22.09\,\std{0.18} & 27.69\,\std{0.60} & 52.97\,\std{1.04} & 33.47\,\std{0.21} & 30.36\,\std{0.23} & 32.39\,\std{0.09} \\
 & \oraclerow{\enspace + Dense-mask} & \oraclerow{24.74\,\ostd{0.33}} & \oraclerow{28.59\,\ostd{0.44}} & \oraclerow{44.37\,\ostd{0.60}} & \oraclerow{35.64\,\ostd{0.72}} & \oraclerow{31.48\,\ostd{2.95}} & \oraclerow{32.41\,\ostd{0.46}} \\
\cmidrule(lr){2-8}
 & CAT-Seg-B                     & 34.51 & 34.52 & 37.82 & 29.95 & 28.95 & 33.12 \\
 
 & \enspace \textbf{+ Ours}        & 36.06\,\std{1.85} & 35.88\,\std{0.60} & 52.66\,\std{2.48} & 43.82\,\std{2.16} & 31.04\,\std{0.46} & 39.67\,\std{0.08} \\
 
 & \oraclerow{\enspace + Dense-mask} & \oraclerow{38.69\,\ostd{0.13}} & \oraclerow{38.22\,\ostd{0.35}} & \oraclerow{60.08\,\ostd{1.37}} & \oraclerow{46.01\,\ostd{1.37}} & \oraclerow{41.61\,\ostd{0.87}} & \oraclerow{44.26\,\ostd{0.49}} \\
\midrule
\multirow{6}{*}{CLIP ViT-L/14}
 & SAN-L                         & 26.23 & 34.51 & 32.00 & 24.15 & 19.24 & 27.40 \\
 & \enspace \textbf{+ Ours}        & 27.43\,\std{0.37} & 37.09\,\std{0.78} & 38.62\,\std{2.69} & 36.13\,\std{0.35} & 31.15\,\std{3.41} & 33.99\,\std{0.90} \\
 
 & \oraclerow{\enspace + Dense-mask} & \oraclerow{31.18\,\ostd{0.31}} & \oraclerow{35.74\,\ostd{0.73}} & \oraclerow{50.05\,\ostd{1.30}} & \oraclerow{38.03\,\ostd{0.26}} & \oraclerow{35.86\,\ostd{0.64}} & \oraclerow{37.64\,\ostd{0.15}} \\
\cmidrule(lr){2-8}
 & CAT-Seg-L                     & 39.36 & 35.64 & 29.52 & 34.22 & 36.41 & 35.26 \\
 
 & \enspace \textbf{+ Ours}        & 42.48\,\std{0.35} & 40.28\,\std{0.35} & 51.83\,\std{0.81} & 52.68\,\std{0.41} & 42.86\,\std{1.45} & 45.88\,\std{0.38} \\

 & \oraclerow{\enspace + Dense-mask} & \oraclerow{44.04\,\ostd{0.41}} & \oraclerow{41.82\,\ostd{1.69}} & \oraclerow{55.48\,\ostd{1.98}} & \oraclerow{49.18\,\ostd{0.69}} & \oraclerow{44.48\,\ostd{0.78}} & \oraclerow{46.67\,\ostd{0.74}} \\
\bottomrule
\end{tabular}%
}
\vspace{-10pt}
\end{table}

\subsection{Main Results}
\label{sec:main_results}

Table~\ref{tab:mess_main} reports mIoU on the MESS benchmark. Our 
method consistently improves over the baseline across all backbones, 
with mean gains of $+7.10$, $+6.55$, $+6.59$, and $+10.62$ mIoU on 
SAN-B, CAT-Seg-B, SAN-L, and CAT-Seg-L, surpassing the supervised 
reference on several specialized domains (e.g., Medical Sciences on 
SAN-B, Engineering on CAT-Seg-L). The improvement is robust to 
backbone scale (ViT-B/16 vs ViT-L/14) and architecture, suggesting 
that prompt disagreement provides a generic source of supervision 
rather than one specific to a particular OVSS design.

Our method is most effective on domains with the largest gap from 
the pretrained distribution. Medical Sciences shows the largest 
improvement on every backbone (e.g., $+19.46$ on SAN-B and $+22.31$ 
on CAT-Seg-L), followed by Engineering and Agriculture \& Biology. 
These results demonstrate the applicability of our method to 
specialized-domain adaptation.

\begin{figure}[t]
\centering
    \includegraphics[width=\linewidth]{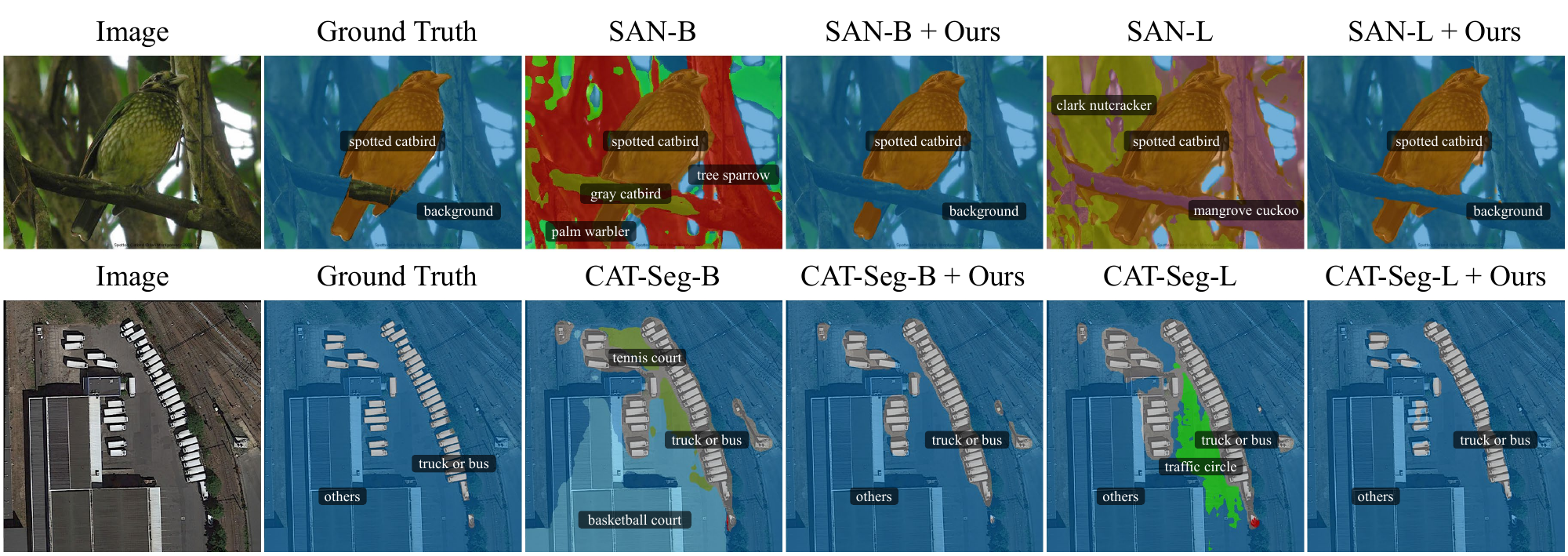}
\caption{Qualitative comparison on the MESS benchmark. Each row shows one 
sample, comparing the zero-shot baseline against our adapted prediction at 
both CLIP scales (ViT-B/16 and ViT-L/14). Top: SAN on CUB-200 (Agriculture \& Biology). Bottom: CAT-Seg on iSAID (Earth Monitoring).}
\label{fig:qualitative}
\vspace{-10pt}
\end{figure}

Figure~\ref{fig:qualitative} shows qualitative comparisons for SAN and CAT-Seg across two CLIP 
scales (ViT-B/16 and ViT-L/14). The baseline zero-shot predictions often miss 
or mislabel domain-specific objects—for example, confusing fine-grained bird 
species in CUB-200 or hallucinating non-existent classes in aerial scenes—
while our adapted predictions align much more closely with the ground truth. 
Improvements are consistent across both scales, indicating that the gains 
observed in Table 1 translate into perceptually clearer segmentations. 
Additional qualitative results spanning all five domain groups for each 
backbone are provided in Appendix~\ref{appendix:additional-qual}.

\begin{figure}[t]
  \centering
  \begin{minipage}[t]{0.48\linewidth}
    \centering
    \captionof{table}{\textbf{Comparison of candidate generation strategies.} We contrast prompt disagreement against MC Dropout and test-time augmentation, with matched candidate size ($K = 14$).}
    \vspace{-3pt}
    \resizebox{\linewidth}{!}{
    \begin{tabular}{l c ccc}
\toprule
 \multirow{2}{*}{Domain}
&  \multirow{2}{*}{Baseline}
& \multicolumn{3}{c}{Strategies} \\
\cmidrule(lr){3-5}
& 
&MC Dropout & TTA & Prompt \\
\midrule
General
& 39.36 & 36.48 & 40.95 & \textbf{42.48} \\
Earth
& 35.64 & 32.10 & 37.73 & \textbf{40.28} \\
Medical
& 29.52 & 42.64 & \textbf{53.15} & 51.83 \\
Engin.
& 34.22 & 47.66 & 50.68 & \textbf{52.68} \\
Agri.\& Bio
& 36.41 & 36.90 & 42.36 & \textbf{42.86} \\
\midrule
Mean
& 35.26 & 39.09 & 44.66 & \textbf{45.88} \\
\bottomrule
\end{tabular}
\label{tab:cand_abl}
    }
  \end{minipage}\hfill
  \begin{minipage}[t]{0.49\linewidth}
    
    \centering
    \captionof{table}{\textbf{Loss function ablation.} 
We ablate the two components of our objective: $\mathcal{L}_{\text{RLPO}}$ 
(preference loss within the queried region) and $\mathcal{L}_{\text{cons}}$ 
(consistency regularizer outside)}
    \vspace{-3pt}
    \resizebox{\linewidth}{!}{
    \begin{tabular}{l c ccc}
\toprule
\multirow{2}{*}{Domain}
& \multirow{2}{*}{Baseline}
& \multicolumn{3}{c}{Ablation} \\
\cmidrule(lr){3-5}
&
& w/o $\mathcal{L}_{\mathrm{RLPO}}$
& w/o $\mathcal{L}_{\mathrm{cons}}$
& \textbf{Ours} \\
\midrule
General
& 39.36 & 41.80 & 41.80 & \textbf{42.48} \\
Earth
& 35.64 & 38.52 & 38.11 & \textbf{40.28} \\
Medical
& 29.52 & 36.25 & 48.93 & \textbf{51.83} \\
Engin.
& 34.22 & 44.83 & 47.88 & \textbf{52.68} \\
Agri.\& Bio
& 36.41 & 39.97 & 41.41 & \textbf{42.86} \\
\midrule
Mean
& 35.26 & 40.51 & 43.45 & \textbf{45.88} \\
\bottomrule
\end{tabular}
\label{tab:loss_abl}
    }
  \end{minipage}
  \vspace{-3pt}
\end{figure}

\begin{table}[t]
\centering
\caption{\textbf{Effect of training set size.}
We vary the number of target-domain training images per dataset from 0, corresponding to the zero-shot baseline, to 128.
We report mIoU for each domain group and the gain over the zero-shot baseline.}
\label{tab:shot_ablation}
\small
\setlength{\tabcolsep}{4pt}
\resizebox{\linewidth}{!}{%
\begin{tabular}{l ccccccc}
\toprule
& \multicolumn{7}{c}{Number of training images} \\
\cmidrule(lr){2-8}
Domain group & $0$ & $4$ & $8$ & $16$ & $32$ & $64$ & $128$ \\
\midrule
General           & 39.36 & 39.57\,\gain{0.21}  & 40.52\,\gain{1.16}  & 40.85\,\gain{1.49}  & 41.96\,\gain{2.60}  & 42.48\,\gain{3.12}  & \textbf{42.56}\,\gain{3.20} \\
Earth Monitoring  & 35.64 & 38.22\,\gain{2.58}  & 37.38\,\gain{1.74}  & 39.50\,\gain{3.86}  & 38.69\,\gain{3.05}  & \textbf{40.28}\,\gain{4.64}  & 39.35\,\gain{3.71} \\
Medical Sciences  & 29.52 & 41.03\,\gain{11.51} & 42.75\,\gain{13.23} & 46.30\,\gain{16.78} & 48.89\,\gain{19.37} & \textbf{51.83}\,\gain{22.31} & 51.42\,\gain{21.90} \\
Engineering       & 34.22 & 36.28\,\gain{2.06}  & 39.19\,\gain{4.97}  & 45.20\,\gain{10.98} & 50.57\,\gain{16.35} & 52.68\,\gain{18.46} & \textbf{54.13}\,\gain{19.91} \\
Agriculture \& Biology & 36.41 & 36.42\,\gain{0.01}  & 38.12\,\gain{1.71}  & 40.16\,\gain{3.75}  & 41.36\,\gain{4.95}  & \textbf{42.86}\,\gain{6.45}  & 41.93\,\gain{5.52} \\
\midrule
{Mean}     & {35.26} & {38.26}\,\gain{3.00} & {39.50}\,\gain{4.24} & {42.31}\,\gain{7.05} & {44.20}\,\gain{8.94} & \textbf{45.88}\,\gain{10.62} & {45.79}\,\gain{10.53} \\
\bottomrule
\end{tabular}%
}
\vspace{-15pt}
\end{table}

\subsection{Ablation Studies}
\label{sec:ablation}

In this section, we ablate three components of our framework: the loss formulation, the candidate source, and the adaptation budget. All ablations are conducted on CAT-Seg-L and follow the default protocol of Section~\ref{sec:main_results}. A hyperparameter sensitivity analysis is provided in Appendix~\ref{appendix:hparam-sensitivity}.

\paragraph{Candidate generation.} 
We use prompt disagreement as the source of comparison candidates for preference queries. Prior visual preference learning methods rely on auxiliary mechanisms such as stochastic sampling, input perturbation, or output sampling to produce candidate diversity~\cite{wallace2024diffusion, liang2025aesthetic,yang2024using,ayupov2025dreamboothdpo}. We instead exploit a source of variation already present in the prompted OVSS interface. This choice has two principled advantages. First, template-induced predictions vary the segmentation hypothesis while preserving the target vocabulary, so candidates remain comparable under the same semantic target—unlike input perturbation, which can shift the visual content itself. Second, no additional mechanism or hyperparameter (e.g., dropout rate, augmentation strength) is introduced; the diversity is obtained for free from the existing prompting interface.

To verify this design empirically, we compare prompt disagreement against 
two representative alternatives: MC Dropout~\cite{gal2016dropout} 
(stochastic forward) and test-time augmentation~\cite{wang2019aleatoric} 
(input perturbation). For MC Dropout, we apply dropout with rate $0.1$ 
during inference. For test-time augmentation, we generate variants of 
each image through horizontal flips and random scaling between $0.75\times$ 
and $1.25\times$. Table~\ref{tab:cand_abl} reports results with a matched 
ensemble size of $K=14$. Prompt disagreement achieves the highest mean 
mIoU ($45.88$), outperforming MC Dropout ($39.09$) and test-time 
augmentation ($44.66$). In Appendix~\ref{appendix:template-analysis}, we further analyze the role of template diversity and observe that combining different types of prompt variation is essential.

\paragraph{Loss components.} 
Our framework combines a region-localized preference loss, \(\mathcal{L}_{\mathrm{RLPO}}\), for supervising the queried region with a consistency regularizer, \(\mathcal{L}_{\mathrm{cons}}\), for stabilizing the loser-template prediction outside that region.
Table~\ref{tab:loss_abl} ablates these two components.
Removing \(\mathcal{L}_{\mathrm{RLPO}}\) reduces the mean mIoU from \(45.88\) to \(40.51\) (\(-5.37\)), whereas removing \(\mathcal{L}_{\mathrm{cons}}\) yields \(43.45\) (\(-2.43\)).
This indicates that RLPO is the primary driver of adaptation, while the consistency regularizer provides complementary stability by mitigating unintended drift outside the queried region.

\paragraph{Number of training images.}
Table~\ref{tab:shot_ablation} reports the effect of varying the number of target images per dataset from $4$ to $128$. The mean mIoU improves rapidly with more images at first, gaining $+3.00$, $+4.24$, and $+7.05$ over the zero-shot baseline at $4$, $8$, and $16$ images, respectively, and continues improving up to $64$ images ($+10.62$). Beyond this point, the mean gain saturates: $128$ images yield $+10.53$, essentially the same as $64$. Different domains exhibit different sample efficiency. On Medical Sciences, our method already gains $+11.51$ mIoU at $4$ images and saturates by $64$ images, while on Engineering the gain grows more gradually and peaks at $128$ images ($+19.91$). These results indicate that preference-guided adaptation is sample-efficient, particularly on specialized domains, where most of the gain is achieved with only a handful of preference queries.

\begin{wrapfigure}{r}{0.5\textwidth}
    \centering
    \vspace{-40pt}
    \includegraphics[width=\linewidth]{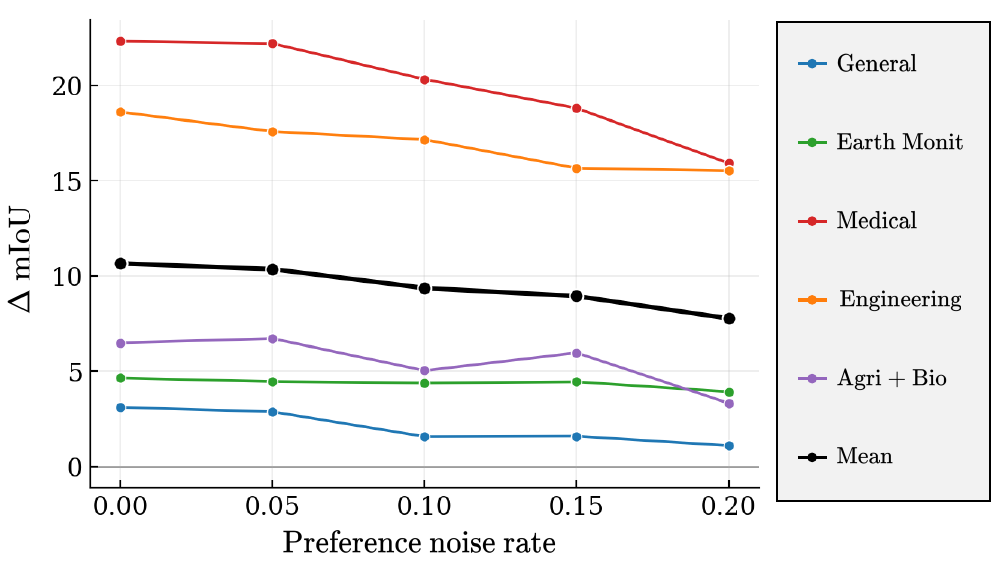}
    \vspace{-15pt}
    \caption{\textbf{Robustness to noisy preferences.} 
We flip a fraction $p$ of preference labels and report the mIoU 
gain over the zero-shot baseline for each domain as well 
as the mean across datasets.}
    \vspace{-15pt}
    \label{fig:noisy}
\end{wrapfigure}
\vspace{-5pt}
\subsection{Robustness Analysis} 
\label{sec:robustness} 
Our method is designed to keep annotation simple by asking the user for only a binary choice between two region-localized candidates. In practice, however, users may still occasionally make mistakes, marking the wrong candidate as preferred. To assess how sensitive our method is to such errors, we simulate noisy oracle feedback: after the oracle selects a winner $w$ and a loser $l$ for each query, we flip the two labels with probability $p$. We sweep $p$ from $0$ (clean) to $0.20$ on CAT-Seg-L and report the mIoU gain over the zero-shot baseline. 

Figure~\ref{fig:noisy} shows the result. Our method is robust to a substantial level of preference noise: at $p = 0.05$ the gain is essentially unchanged from the clean setting, and even at $p = 0.20$ the method still recovers roughly $8$ mIoU on average. Medical Sciences and Engineering, which benefit most from adaptation, also retain the largest gains under heavy noise, indicating that the robustness extends to the regimes where our method is most useful. This robustness suggests that our framework can tolerate the imperfect feedback expected in real world applications.


\vspace{-5pt}
\section{Conclusion}
\label{sec:conclusion}
\vspace{-7pt}
We presented a preference-guided adaptation framework for open-vocabulary semantic segmentation that turns prompt disagreement into an actionable supervision signal. By mining localized comparison queries from prompt ensembles and learning from them via Region-Localized Preference Optimization with consistency regularization, our method adapts OVSS models to specialized target domains using only binary preferences over small image regions. Across the MESS benchmark, we observed consistent improvements without dense masks. The method remains sample-efficient at small adaptation budgets and robust under noisy preference feedback, suggesting that prompt-induced variation can serve as a practical supervision signal for OVSS adaptation in regimes where dense annotation is costly.
\vspace{-8pt}
\section{Limitations}
\label{sec:limitations}
\vspace{-7pt}

Our framework assumes that the target vocabulary can be reasonably expressed through natural-language prompts and that prompt-induced candidates contain at least one useful hypothesis within the queried region. When all templates produce similarly inaccurate predictions, the preference signal becomes less informative, since selecting the “less wrong” candidate provides limited guidance toward the correct segmentation. This limitation is likely to arise for highly domain-specific concepts whose terminology, visual appearance, or label granularity is not well captured by natural-image-style templates. Extending preference-guided adaptation with richer prompt sources, such as learned, domain-specific, or expert-provided templates, is a promising direction for future work.

\medskip

\bibliographystyle{unsrtnat}
\bibliography{bib}

\clearpage
\appendix
\section{Derivation of the Region-Localized Preference Optimization Loss}
\label{app:rlpo}
 
We derive the RLPO loss by specializing the standard DPO framework \cite{rafailov2023direct} to OVSS. Section~\ref{app:rlpo-recap} compresses the standard derivation; Section~\ref{app:rlpo-adapt} carries out the specialization, including the cross-template Bradley--Terry comparison, the class-balanced regional score, and the online-DPO interpretation.
 
\subsection{DPO Recap}
\label{app:rlpo-recap}
 
We briefly recall the standard DPO derivation \cite{rafailov2023direct}. Given a generative policy $\pi(y \mid c)$ over responses $y \in \mathcal{Y}$ conditioned on a context $c$, the KL-regularized objective
\begin{equation}
\max_{\pi}\;
\mathbb{E}_{c \sim \mathcal{D}}\!\left[
\mathbb{E}_{y \sim \pi(\cdot \mid c)}[\,r(c,y)\,]
\;-\; \beta\, D_{\mathrm{KL}}\!\left(\pi(\cdot \mid c) \,\|\, \pi_{\mathrm{ref}}(\cdot \mid c)\right)
\right]
\label{eq:kl-rl}
\end{equation}
admits the closed-form per-context optimum
\begin{equation}
\pi^{*}(y \mid c)
= \frac{1}{Z(c)}\, \pi_{\mathrm{ref}}(y \mid c) \exp\!\left(\tfrac{1}{\beta} r(c,y)\right),
\label{eq:closed-form}
\end{equation}
which can be equivalently rearranged as a log-ratio expression for the reward,
\begin{equation}
r(c,y) = \beta \log \frac{\pi^{*}(y \mid c)}{\pi_{\mathrm{ref}}(y \mid c)} + \beta \log Z(c).
\label{eq:reward-logratio}
\end{equation}
Substituting Eq.~\eqref{eq:reward-logratio} into the Bradley--Terry preference $P(y_w \succ y_l \mid c) = \sigma(r(c, y_w) - r(c, y_l))$ \cite{bradley1952rank}, the context-dependent normalizers $\beta \log Z(c)$ cancel, yielding the DPO loss
\begin{equation}
\mathcal{L}_{\mathrm{DPO}}(\theta)
= - \mathbb{E}_{(c, y_w, y_l) \sim \mathcal{D}_{\mathrm{pref}}}\!\left[
\log \sigma\!\left(
\beta \log \frac{\pi_\theta(y_w \mid c)}{\pi_{\mathrm{ref}}(y_w \mid c)}
- \beta \log \frac{\pi_\theta(y_l \mid c)}{\pi_{\mathrm{ref}}(y_l \mid c)}
\right)
\right].
\label{eq:dpo}
\end{equation}
The remainder of this appendix specializes Eq.~\eqref{eq:dpo} to our setting.
 
\subsection{Adaptation to Template-Conditional Segmentation}
\label{app:rlpo-adapt}
 
\paragraph{Setting.}
We instantiate the abstract context $c$ of Section~\ref{app:rlpo-recap} as $c = (x, t_k)$, an input image $x \in \mathbb{R}^{H \times W \times 3}$ paired with one of $K$ prompt templates $t_k \in \mathcal{T}$; the term \emph{prompt} hereafter refers exclusively to these OVSS templates, while \emph{context} denotes the DPO input as in Section~\ref{app:rlpo-recap}. The response of template $k$ is the segmentation it induces:
\begin{equation}
\hat Y^k = \bigl\{ \hat Y^k(\mathbf{u}) \bigr\}_{\mathbf{u} \in \Omega},
\qquad
\hat Y^k(\mathbf{u}) = \arg\max_{c' \in \mathcal{C}} P_\theta^k(c' \mid x, \mathbf{u}).
\end{equation}
Under a pixelwise-independence factorization $\pi_\theta(\hat Y^k \mid x, t_k) = \prod_{\mathbf{u} \in \Omega} P_\theta^k(\hat Y^k(\mathbf{u}) \mid x, \mathbf{u})$, which is standard for OVSS heads producing per-pixel softmax outputs, the response log-likelihood decomposes additively across pixels,
\begin{equation}
\log \pi_\theta(\hat Y^k \mid x, t_k)
= \sum_{\mathbf{u} \in \Omega} \log \max_{c' \in \mathcal{C}} P_\theta^k(c' \mid x, \mathbf{u}),
\label{eq:loglik-pixel}
\end{equation}
where $\log P_\theta^k(\hat Y^k(\mathbf{u}) \mid x, \mathbf{u}) = \log \max_{c'} P_\theta^k(c' \mid x, \mathbf{u})$ holds by the definition of $\hat Y^k(\mathbf{u})$.
 
\paragraph{Cross-template Bradley--Terry comparison.}
Standard DPO compares two responses under a \emph{shared} context, so that the context-dependent normalizers $\beta \log Z(c)$ in Eq.~\eqref{eq:reward-logratio} cancel inside the BT difference. In our setting the winner and loser responses arise from \emph{sibling} contexts $c_w = (x, t_w)$ and $c_l = (x, t_l)$ that share the underlying image $x$ and the target vocabulary $\mathcal{C}$ but differ in the prompt template. Substituting Eq.~\eqref{eq:reward-logratio} into the BT preference therefore yields
\begin{equation}
\begin{aligned}
&r(c_w, \hat Y^w) - r(c_l, \hat Y^l) \\
&\quad = \beta \!\left[ \log\frac{\pi_\theta(\hat Y^w \mid c_w)}{\pi_{\mathrm{ref}}(\hat Y^w \mid c_w)} - \log\frac{\pi_\theta(\hat Y^l \mid c_l)}{\pi_{\mathrm{ref}}(\hat Y^l \mid c_l)} \right] \\
&\qquad + \underbrace{\beta\!\left[\log Z(c_w) - \log Z(c_l)\right]}_{=:\, \Delta_Z(c_w, c_l),\ \text{constant in } \theta}.
\end{aligned}
\label{eq:cross-prompt-bt}
\end{equation}
The residual term $\Delta_Z(c_w, c_l)$ does not vanish in general, but it depends only on the ground-truth reward and the reference policy, not on the trainable parameters $\theta$. As an additive offset inside the BT log-sigmoid loss it contributes no gradient to the optimization, so the closed form of Eq.~\eqref{eq:dpo} extends to our cross-template setting up to a $\theta$-independent constant. The KL regularization of Eq.~\eqref{eq:kl-rl} is thus inherited per template: each context $c_k = (x, t_k)$ is anchored to its own $\pi_{\mathrm{ref}}(\cdot \mid c_k)$, which is precisely the regularization our adaptation needs.
 
\paragraph{Region restriction.}
Preferences in our setting are elicited over a query region $R \subseteq \Omega$ rather than the full image. Restricting the sum in Eq.~\eqref{eq:loglik-pixel} to $R$ yields a region-localized log-likelihood,
\begin{equation}
\bar S_\theta^k(R)
= \sum_{\mathbf{u} \in R} \log \max_{c' \in \mathcal{C}} P_\theta^k(c' \mid x, \mathbf{u}),
\label{eq:score-region}
\end{equation}
which is the strict consequence of Eq.~\eqref{eq:loglik-pixel} under region restriction.
 
\paragraph{Class-balanced averaging.}
Eq.~\eqref{eq:score-region} weights each pixel uniformly, so a class that occupies most of $R$ dominates the score and can suppress contributions from smaller but semantically important classes. To prevent this, we replace the uniform sum with a class-balanced average that equalizes contributions across present classes:
\begin{equation}
S_\theta^k(R)
= \frac{1}{|\mathcal{U}_R|}
\sum_{c \in \mathcal{U}_R}
\frac{1}{|R_c|}
\sum_{\mathbf{u} \in R_c}
\log \max_{c' \in \mathcal{C}} P_\theta^k(c' \mid x, \mathbf{u}),
\label{eq:score-app}
\end{equation}
with buckets $R_c = \{ \mathbf{u} \in R : \hat Y^w(\mathbf{u}) = c \}$ and present-class set $\mathcal{U}_R = \{ c \in \mathcal{C} : R_c \neq \emptyset \}$. We use the winner's hard prediction $\hat Y^w$ to define a stable, $k$-independent partition: the same buckets $\{R_c\}$ are used in all four score evaluations $\{(\theta, w), (\theta, l), (\mathrm{ref}, w), (\mathrm{ref}, l)\}$, so the BT difference depends only on the per-template logits, not on which template's prediction is used to bucket. We view this class-balanced averaging as a deliberate departure from the strict derivation, motivated by class imbalance within $R$, rather than a derived consequence of Eq.~\eqref{eq:dpo}. The same construction defines $S_{\mathrm{ref}}^k(R)$ by replacing $P_\theta^k$ with $P_{\mathrm{ref}}^k$.
 
\paragraph{Online-DPO interpretation.}
Within each optimization step, $\hat Y^k$ is a fixed discrete label, so the BT closed form of Eq.~\eqref{eq:dpo} applies to $(y_w, y_l) = (\hat Y^w, \hat Y^l)$. Across steps, $\hat Y^k$ evolves with $\theta$, placing RLPO within the family of online/iterative DPO methods \cite{calandriello2024human, xiong2023iterative}; the $\arg\max$ corresponds to a deterministic mode response equivalent to the zero-temperature limit of softmax sampling.
 
\paragraph{The RLPO loss.}
Combining the cross-template BT extension with the class-balanced region score, and defining the region-localized implicit reward $r^k(R) = \beta\!\left[ S_\theta^k(R) - S_{\mathrm{ref}}^k(R) \right]$, the BT closed form of Eq.~\eqref{eq:dpo} applied to $(\hat Y^w, \hat Y^l)$ yields
\begin{equation}
\begin{aligned}
\mathcal{L}_{\mathrm{RLPO}}(\theta)
&= -\log \sigma\!\left( r^w(R) - r^l(R) \right) \\
&= -\log \sigma\!\left( \beta\!\left[ \bigl( S_\theta^w(R) - S_{\mathrm{ref}}^w(R) \bigr) - \bigl( S_\theta^l(R) - S_{\mathrm{ref}}^l(R) \bigr) \right] \right).
\end{aligned}
\label{eq:rlpo-app}
\end{equation}
 
 

\section{Template Analysis}
\label{appendix:template-analysis}

We further analyze how different prompt-template variations affect adaptation.
Table~\ref{tab:vild-templates} lists the full ViLD template pool used in our main experiments.
Table~\ref{tab:template-comp} compares the full ViLD pool with two diagnostic template sets---a sentence-only ViLD subset and a controlled scale-only set---to isolate the roles of sentence and scale variation. The \emph{sentence-only} subset consists of the five ViLD templates without scale modifiers (rows 1--5).
The \emph{scale-only} subset fixes the sentence structure to
\texttt{a photo of a [scale] \{\} in the scene}
and varies only the scale modifier over
\{\texttt{none}, \texttt{small}, \texttt{medium}, \texttt{large}\}.

The sentence-only and scale-only subsets achieve similar mean mIoU, \(43.79\) and \(43.65\), respectively, suggesting that scale modifiers alone do not consistently improve segmentation quality.
However, their effects are complementary across domains: scale-only templates improve over sentence-only templates in some agriculture and engineering datasets, whereas sentence-only templates are stronger in several medical datasets.
Using the full 14-template set achieves the best mean mIoU of \(45.88\), outperforming both subsets.
\begin{table}[t]
\centering
\caption{Per-dataset adaptation performance for three template subsets that vary along different diversity axes. The best per dataset across the three subsets is highlighted in 
bold; the zero-shot baseline is shown for reference.}
\label{tab:template-comp}
\small
\setlength{\tabcolsep}{3pt}
\resizebox{\linewidth}{!}{%
\begin{tabular}{l cccc cccc ccc cccc ccc| c}
\toprule
& \multicolumn{4}{c}{General} & \multicolumn{4}{c}{Earth Monit.} & \multicolumn{3}{c}{Medical} & \multicolumn{4}{c}{Engineering} & \multicolumn{3}{c}{Agri.\ \& Bio.} & \\
\cmidrule(lr){2-5} \cmidrule(lr){6-9} \cmidrule(lr){10-12} \cmidrule(lr){13-16} \cmidrule(lr){17-19}
Subset
& \rotatebox{90}{BDD100K} & \rotatebox{90}{MHP v1} & \rotatebox{90}{FoodSeg103} & \rotatebox{90}{ATLANTIS}
& \rotatebox{90}{iSAID} & \rotatebox{90}{WorldFloods} & \rotatebox{90}{FloodNet} & \rotatebox{90}{UAVid}
& \rotatebox{90}{Kvasir-Inst.} & \rotatebox{90}{CHASE\_DB1} & \rotatebox{90}{PAXRay-4}
& \rotatebox{90}{Corrosion CS} & \rotatebox{90}{DeepCrack} & \rotatebox{90}{PST900} & \rotatebox{90}{ZeroWaste-f}
& \rotatebox{90}{SUIM} & \rotatebox{90}{CUB-200} & \rotatebox{90}{CWFID}
& \textit{Mean} \\
\midrule
Baseline       & 48.23 & 30.77 & 32.92 & 45.51 & 19.67 & 39.94 & 41.05 & 41.90 & 65.49 & 3.32 & 19.75 & 7.47 & 25.27 & 78.73 & 25.42 & 49.75 & 21.89 & 37.58 & \textit{35.26} \\
\midrule
Sentence-only  & 50.76 & \textbf{36.95} & 36.73 & \textbf{45.70} & 35.74 & 39.90 & \textbf{41.19} & 43.76 & 80.21 & 19.65 & 46.05 & 21.40 & 60.22 & 80.75 & 24.33 & 55.74 & 22.38 & 46.69 & \textit{43.79} \\
Scale-only     & \textbf{50.89} & 34.67 & \textbf{37.41} & 45.41 & 32.59 & 40.33 & 41.16 & \textbf{46.19} & 76.12 & 20.14 & 47.09 & 20.94 & 61.44 & 79.98 & 24.76 & 57.71 & \textbf{22.47} & 46.48 & \textit{43.65} \\
Full pool      & 50.58 & 36.41 & \textbf{37.41} & 45.50 & \textbf{36.17} & \textbf{41.13} & 41.07 & 42.74 & \textbf{87.41} & \textbf{20.74} & \textbf{47.34} & \textbf{22.90} & \textbf{74.39} & \textbf{80.96} & \textbf{32.47} & \textbf{58.49} & 22.22 & \textbf{47.87} & \textit{\textbf{45.88}} \\
\bottomrule
\end{tabular}%
}
\end{table}
This indicates that the benefit of scale-aware prompts is not simply higher standalone accuracy, but the additional candidate diversity they introduce around the same target vocabulary.
By combining sentence and scale variations, the full prompt ensemble provides more diverse yet semantically aligned segmentation hypotheses, which improves preference-query mining.

\begin{table}[t]
  \centering
  \vspace{-10pt}
  \caption{Detailed Implementation configuration.}
  \label{tab:appendix-config}
  \resizebox{0.8\linewidth}{!}{

  \begin{minipage}[t]{0.49\linewidth}
    \centering
    \vspace{3pt}
    \resizebox{\linewidth}{!}{
    \begin{tabular}{ll}
\toprule
\multicolumn{2}{c}{CAT-Seg~\cite{cho2024cat}}\\
\midrule
optimizer & AdamW \\
weight decay & 1e-4 \\
learning rate & 3e-3 \\
$\beta$ & 0.1 \\
$\lambda_{\text{cons}}$ & 0.1 \\
$\tau_{\text{conf}}$ & 0.8 \\
$q$ & 0.95 \\
vision LoRA rank, $\alpha$ & $r=4$, $\alpha=1.0$ \\
text adapter rank & $r=4$ \\
exception lr   & 1e-3$^\dagger$ \\
\bottomrule
\end{tabular}
    }
    {
    \vspace{5pt}
    \footnotesize $^\dagger$Applied to \texttt{cub\_200}, \texttt{atlantis}, \texttt{isaid}, \texttt{pst900}.}

  \end{minipage}\hfill
  \begin{minipage}[t]{0.49\linewidth}
    \centering
    \vspace{3pt}
    \resizebox{\linewidth}{!}{
    \begin{tabular}{ll}
\toprule
\multicolumn{2}{c}{SAN~\cite{xu2023side}}\\
\midrule
optimizer & AdamW \\
weight decay & 1e-4 \\
learning rate & 1e-3 \\
$\beta$ & 0.1 \\
$\lambda_{\text{cons}}$ & 0.1 \\
$\tau_{\text{conf}}$ & 0.8 \\
$q$ & 0.95 \\
vision LoRA rank, $\alpha$ & $r=4$, $\alpha=1.0$ \\
text adapter rank & $r=4$ \\
exception lr & 3e-4$^\dagger$ \\
\bottomrule
\end{tabular}
    }
    \vspace{5pt}
    {\footnotesize $^\dagger$Applied to \texttt{cub\_200}, \texttt{atlantis}, \texttt{pst900}.}

  \end{minipage}
  }
  \vspace{-10pt}
  
\end{table}

\section{Implementation Details}
\label{appendix:implementation}

\paragraph{Adapter configuration.}
The adapter consists of two lightweight modules: a vision LoRA on the 
CLIP image encoder and a residual adapter on the text classifier 
embedding. Both modules are rank~4. The vision LoRA is attached to the 
Q and V projections of the last four transformer blocks of the CLIP 
image encoder. The text adapter is a low-rank residual on the 
mean-of-$K$ classifier embedding. All other backbone parameters are 
kept frozen.

\paragraph{Training protocol.}
We adapt each OVSS backbone with AdamW (weight decay 1e-4) and a 
constant learning rate, kept fixed throughout the streaming run. Across 
all backbones we use the same DPO temperature $\beta = 0.1$, consistency 
weight $\lambda_{\text{cons}} = 0.1$, confidence threshold 
$\tau_{\text{conf}} = 0.8$, and entropy quantile $q = 0.95$. Each 
adaptation step processes a single image (batch size~1), consistent 
with the streaming protocol of Section~\ref{sec:problem-setting}.
The default learning rate differs by backbone. For a small 
subset of datasets, we additionally 
use a lower learning rate. The full configuration, including 
backbone-specific learning rates and dataset-specific exceptions, is 
summarized in Table~\ref{tab:appendix-config}.

\section{Validation of the Preference Oracle}
\label{appendix:oracle-validation}
 
Throughout the main experiments, each binary preference is supplied by an oracle
that compares the two candidate predictions against the ground-truth mask within the
queried region (Section~\ref{sec:setup}). While this enables large-scale evaluation,
it raises the question of how well the oracle reflects human judgments. We examine
this in two ways: whether humans agree with the oracle
(Appendix~\ref{appendix:oracle-agreement}) and whether the method remains robust to
errors concentrated on cases where humans and the oracle disagree
(Appendix~\ref{appendix:oracle-noise}).

\subsection{Human--Oracle Agreement}
\label{appendix:oracle-agreement}
 
\begin{table}[t]
\centering
\small
\setlength{\tabcolsep}{6pt}
\renewcommand{\arraystretch}{1.15}
\caption{\textbf{Agreement between human annotators and the preference
oracle.} Mined preference pairs are split into three difficulty tiers by the IoU
margin between the two candidates, with boundaries set to the terciles of the
margin distribution observed during actual adaptation runs. Each of the $20$
participants judged $30$ pairs ($10$ per tier), for $600$ judgments in total.}
\vspace{3pt}
\label{tab:human-agreement}
\begin{tabular}{lcccc}
\toprule
\textbf{Tier} & IoU margin & \#Judgments & \textbf{Agreement} \\
\midrule
Easy    & $\geq 0.18$      & 200  & 0.967 \\
Medium  & $[0.04,\ 0.18)$  & 200  & 0.953 \\
Hard    & $< 0.04$         & 200 & 0.840 \\
\midrule
\textbf{Overall} & ---     & 600 & \textbf{0.920} \\
\bottomrule
\end{tabular}
\end{table}

\paragraph{Protocol.}
We construct a preference-pair database by running Preference Query Mining
(Section~\ref{sec:query}), the same procedure that generates comparisons
during adaptation, so that the pairs shown to participants follow the same
distribution as those encountered in an actual run. Pairs are grouped into
\emph{easy}, \emph{medium}, and \emph{hard} tiers by the IoU margin between the two
candidates inside the queried region. The tier boundaries, $0.04$ and $0.18$, are
set to the terciles of the margin distribution over all queries answered during our
adaptation runs, so each tier reflects a third of the comparisons the method
actually issues. Each participant judged $30$ pairs, $10$ sampled from each tier,
which allows agreement to be estimated separately at every difficulty level rather
than only in aggregate. In total, $20$ participants provided $600$ judgments.
 
\paragraph{Results.}
Table~\ref{tab:human-agreement} reports agreement with the oracle. Humans select
the same candidate as the oracle on $0.920$ of all pairs, taking $9.4$ seconds per
judgment on average. Agreement is near-ceiling on easy and medium pairs ($0.967$
and $0.953$) and drops on hard pairs ($0.840$), which is the expected pattern: when
the two candidates are separated by a very small IoU margin they are close to
equally good, and the choice becomes genuinely ambiguous rather than incorrect.
The oracle is therefore a close proxy for human preference over the range of
comparisons our method issues, with the residual disagreement concentrated in
near-tie queries.
 
 
 
 

\subsection{Noise Concentrated on Near-Tie Queries}
\label{appendix:oracle-noise}
 
\begin{table}[t]
\centering
\small
\setlength{\tabcolsep}{5pt}
\renewcommand{\arraystretch}{1.15}
\caption{\textbf{Robustness to preference noise placed where humans actually
disagree} (CAT-Seg-L, mIoU~\%). Noise is injected only into near-tie queries
whose IoU margin falls below $\tau$. \textit{Random} replaces the oracle choice
with a coin flip; \textit{Flip} actively inverts it, an adversarial upper bound
on annotator error.}
\vspace{3pt}
\label{tab:margin-noise}
\resizebox{\linewidth}{!}{%
\begin{tabular}{llcccccc}
\toprule
\textbf{Condition} & $\tau$ & General & Earth Monit. & Medical Sci. & Engineering & Agri.\ \& Biology & \textbf{Mean} \\
\midrule
Zero-shot          & ---    & 39.36 & 35.64 & 29.52 & 34.22 & 36.41 & 35.26 \\
Clean (oracle)     & ---    & 42.48 & 40.28 & 51.83 & 52.68 & 42.86 & \textbf{45.88} \\
\midrule
\multirow{3}{*}{Random}
                   & 0.05   & 42.39 & 39.83 & 51.60 & 49.38 & 43.74 & 45.13 \\
                   & 0.10   & 42.08 & 39.18 & 50.66 & 51.95 & 40.85 & 44.85 \\
                   & 0.15   & 42.21 & 37.52 & 50.57 & 48.68 & 42.93 & 44.12 \\
\midrule
\multirow{3}{*}{Flip}
                   & 0.05   & 42.74 & 39.25 & 51.30 & 51.78 & 41.62 & 45.21 \\
                   & 0.10   & 41.62 & 38.93 & 48.86 & 48.40 & 37.33 & 43.02 \\
                   & 0.15   & 40.80 & 39.32 & 46.93 & 49.10 & 35.34 & 42.43 \\
\bottomrule
\end{tabular}%
}
\vspace{-10pt}
\end{table}

Section~\ref{sec:robustness} studies robustness under preference labels flipped
uniformly at random. The agreement study above suggests a more targeted stress
test, since human error is not uniform but concentrated on hard, near-tie pairs.
We therefore inject noise only into queries whose IoU margin falls below a
threshold $\tau$, under two models: \emph{Random} replaces the oracle choice with a
coin flip, and \emph{Flip} actively inverts it, which is adversarial rather than
merely noisy and upper-bounds any realistic annotator.
 
Table~\ref{tab:margin-noise} reports the result. Performance degrades gracefully in
both models. Even in the most severe setting, where every near-tie query with
margin below $0.15$ is actively inverted, the mean remains at $42.43$, well above
the $35.26$ zero-shot baseline. Since humans disagree with the oracle on only $16\%$
of hard pairs, the realistic error regime sits comfortably inside the range the
method tolerates.

\section{Comparison with Alternative Adaptation Strategies}
\label{appendix:alternatives}

Section~\ref{sec:setup} compares our method with the zero-shot baseline and a
matched-budget Dense-mask reference while keeping the adaptation protocol fixed.
Because no prior method uses an equivalent preference-supervision budget, we further
compare against alternatives ranging from no annotation to stronger point and dense
supervision. All methods use the same CAT-Seg-L backbone and the same $64$
target-domain images, with results summarized in Table~\ref{tab:alternatives}.

\paragraph{Strategies compared.}
\textit{Prompt ensemble} averages predictions from the $K$ templates in our candidate
pool without training or annotation.
\textit{Weakly-sup.\ (point)} supervises one point at the distance-transform maximum
of each connected component, requiring instance-level localization.
\textit{Prompt selection} uses dense ground-truth masks to select the best template
for each dataset, without updating model parameters.
\textit{Dense-mask}, introduced in Section~\ref{sec:setup}, follows our single-step
adaptation protocol but replaces the binary preference with a ground-truth mask.
\textit{Supervised} is a fully supervised upper reference, using GT-based prompt
tuning following ~\cite{zhou2022learning} and ~\cite{rao2022denseclip}
on dense masks from the same $64$ images for $200$ epochs. Together, these baselines
span supervision from annotation-free inference to full dense supervision.

\providecommand{\tbd}{\textcolor{red}{--}}

\begin{table}[t]
\centering
\small
\setlength{\tabcolsep}{4pt}
\renewcommand{\arraystretch}{1.15}
\caption{\textbf{Comparison with alternative adaptation strategies on CAT-Seg-L}
(mIoU, \%).
\textbf{Annotation} denotes target-domain supervision as \textit{type / scope}, with
rows ordered by annotation cost. \textcolor{oracle}{\textit{Grayed}} rows require
stronger supervision than binary preferences and are included as references. Results are mean $\pm$ standard deviation over three runs.}
\vspace{3pt}
\label{tab:alternatives}
\resizebox{\linewidth}{!}{%
\begin{tabular}{llcccccc}
\toprule
\textbf{Method} & \textbf{Annotation} & General & Earth Monit. & Medical Sci. & Engineering & Agri.\ \& Biology & \textbf{Mean} \\
\midrule
CAT-Seg-L & --- & 39.36 & 35.64 & 29.52 & 34.22 & 36.41 & 35.26 \\
\enspace + Prompt ens. & --- & 39.53 & 37.66 & 25.11 & 34.97 & 33.78 & 34.74 \\
\enspace \textbf{+ Ours} & binary preference / image & 42.48\,\std{0.35} & 40.28\,\std{0.35} & 51.83\,\std{0.81} & 52.68\,\std{0.41} & 42.86\,\std{1.45} & 45.88\,\std{0.38} \\
\cmidrule(lr){1-8}
\oraclerow{\enspace + Weakly-sup.\ (point)} & \oraclerow{click / object} & \oraclerow{41.79\,\std{0.62}} & \oraclerow{34.20\,\std{0.86}} & \oraclerow{51.86\,\std{2.70}} & \oraclerow{44.15\,\std{2.07}} & \oraclerow{42.43\,\std{0.97}} & \oraclerow{42.41\,\std{0.68}} \\
\oraclerow{\enspace + Prompt selection} & \oraclerow{dense mask / image} & \oraclerow{39.61\,\std{0.29}} & \oraclerow{40.15\,\std{0.73}} & \oraclerow{35.05\,\std{0.10}} & \oraclerow{39.13\,\std{0.04}} & \oraclerow{35.67\,\std{0.26}} & \oraclerow{38.20\,\std{0.08}} \\
\oraclerow{\enspace + Dense-mask} & \oraclerow{dense mask / image} & \oraclerow{44.04\,\std{0.41}} & \oraclerow{41.82\,\std{1.69}} & \oraclerow{55.48\,\std{1.98}} & \oraclerow{49.18\,\std{0.69}} & \oraclerow{44.48\,\std{0.78}} & \oraclerow{46.67\,\std{0.74}} \\
\oraclerow{\enspace + Supervised} & \oraclerow{dense mask / image} & \oraclerow{45.83\,\std{0.23}} & \oraclerow{46.96\,\std{1.36}} & \oraclerow{73.03\,\std{0.50}} & \oraclerow{55.91\,\std{0.86}} & \oraclerow{47.73\,\std{0.51}} & \oraclerow{53.17\,\std{0.50}} \\
\bottomrule
\end{tabular}%
}
\vspace{-10pt}
\end{table}

\paragraph{Results.}
Prompt ensembling does not improve over the zero-shot baseline ($34.74$ versus
$35.26$). Prompt selection reaches
$38.20$ despite requiring per-dataset dense masks, so template choice alone does not
account for the gain, and the point baseline reaches $42.41$ while requiring the
annotator to click every object instance. Our method surpasses both at $45.88$ with a
single binary judgment per image. The two mask-supervised references bracket it from
above: at a matched single-step budget dense masks reach $46.67$, and multi-epoch
fully-supervised prompt tuning reaches $53.17$, which we do not claim to match.

\section{Detailed Results on MESS Benchmark}
\label{appendix:per-dataset}

We provide details of the MESS datasets used in this work and 
report per-dataset performance for each of the four base OVSS 
backbones, as well as per-dataset breakdowns of the ablation studies 
in Section~\ref{sec:ablation}. Table~\ref{tab:dataset-detail} lists 
the datasets grouped by domain, along with class count, license, and 
a sample of class labels.

Table~\ref{tab:per-dataset-main} expands the group-level numbers in 
Table~\ref{tab:mess_main} of the main paper with per-dataset results 
for all four backbones (SAN-B, CAT-Seg-B, SAN-L, and CAT-Seg-L). 
Within each backbone, the better of the zero-shot baseline and our 
method is highlighted in bold; the supervised reference is shown for 
reference and excluded from the comparison.

Tables~\ref{tab:per-dataset-loss}--\ref{tab:per-dataset-shot} expand 
the three ablation studies of Section~\ref{sec:ablation} with 
per-dataset results on CAT-Seg-L. Table~\ref{tab:per-dataset-loss} 
breaks down the loss ablation of 
Table~\ref{tab:loss_abl}, comparing the full method against 
ablating $\mathcal{L}_{\text{RLPO}}$ and $\mathcal{L}_{\text{cons}}$. 
Table~\ref{tab:per-dataset-candidate} contrasts prompt disagreement 
against MC Dropout and test-time augmentation under matched candidate 
size, expanding Table~\ref{tab:cand_abl}. Finally, 
Table~\ref{tab:per-dataset-shot} reports the effect of varying the 
number of target-domain training images per dataset from 0 (zero-shot 
baseline) to 128, expanding Table~\ref{tab:shot_ablation}. The best 
per dataset is highlighted in bold across all three ablation tables.

\section{Hyperparameter Sensitivity}
\label{appendix:hparam-sensitivity}

We analyze the sensitivity of our method to four hyperparameters on CAT-Seg-L: the DPO temperature \(\beta\), the consistency weight \(\lambda_{\mathrm{cons}}\), the entropy quantile \(q\) used for query-region selection, and the confidence threshold \(\tau_{\mathrm{conf}}\) for winner-pseudo-label pixels.
Each ablation varies one hyperparameter while keeping the others fixed to their default values.
Table~\ref{tab:hyper} reports mIoU for each domain group and the mean across groups.
Across all four hyperparameters, the mean mIoU remains within approximately \(\pm 1\) point of the default setting, and every variation stays substantially above the zero-shot baseline of \(35.26\).
The default values achieve the best overall mean and are competitive across domain groups, indicating that our method is robust to hyperparameter choices.

\begin{table}[t]
\centering
\caption{Sensitivity to the hyperparameters of our method. Default values and the best per domain group within each hyperparameter group are shown in bold.}
\vspace{5pt}
\label{tab:hyper}
\small
\setlength{\tabcolsep}{5pt}
\begin{tabular}{c|l| ccccc|c}
\toprule
Param & Value & General & Earth Monit. & Medical Sci. & Engineering & Agri.\ \& Bio. & Mean \\
\midrule
\multirow{3}{*}{$\beta$}
& 0.05         & \textbf{42.50} & 39.95          & \textbf{52.02} & 50.27          & 43.01          & \textit{45.33} \\
& \textbf{0.1} & 42.47          & 40.28          & 51.83          & \textbf{52.68} & 42.86          & \textit{\textbf{45.88}} \\
& 0.2          & 41.77          & \textbf{40.31} & 51.61          & 51.84          & \textbf{43.24} & \textit{45.57} \\
\midrule
\multirow{3}{*}{$\lambda_{\text{cons}}$}
& 0.05         & 41.60          & 40.21          & 50.98          & 52.06          & 42.42          & \textit{45.31} \\
& \textbf{0.1} & \textbf{42.47} & \textbf{40.28} & 51.83          & \textbf{52.68} & \textbf{42.86} & \textit{\textbf{45.88}} \\
& 0.2          & 41.90          & 39.83          & \textbf{51.96} & 50.41          & 41.80          & \textit{44.99} \\
\midrule
\multirow{3}{*}{$q$}
& 0.90          & 42.31          & 39.84          & 50.00          & \textbf{53.80} & 41.53          & \textit{45.47} \\
& \textbf{0.95} & \textbf{42.47} & \textbf{40.28} & 51.83          & 52.68          & \textbf{42.86} & \textit{\textbf{45.88}} \\
& 0.98          & 41.70          & 39.30          & \textbf{52.65} & 52.48          & 41.95          & \textit{45.43} \\
\midrule
\multirow{3}{*}{$\tau_{\text{conf}}$}
& 0.70          & 41.95          & 40.02          & 51.39          & 51.25          & \textbf{43.43} & \textit{45.41} \\
& \textbf{0.80} & \textbf{42.47} & 40.28          & \textbf{51.83} & \textbf{52.68} & 42.86          & \textit{\textbf{45.88}} \\
& 0.90          & 41.82          & \textbf{40.68} & 51.29          & 50.34          & 42.72          & \textit{45.19} \\
\bottomrule
\end{tabular}
\end{table}

\begin{table}[t]
\centering
\vspace{-15pt}
\caption{Compute and memory profile of our adaptation on single NVIDIA H200 GPU.}
\vspace{5pt}

\label{tab:efficiency}
\small
\begin{tabular}{lrr}
\toprule
Metric & CAT-Seg-L & SAN-L \\
\midrule
Full model parameters             & 433.7\,M & 436.7\,M \\
Trainable parameters              & 71{,}681 & 71{,}681 \\
\quad Vision LoRA                 & 65{,}536 & 65{,}536 \\
\quad Text residual adapter       &  6{,}145 &  6{,}145 \\
Trainable fraction                & 0.017$\,\%$ & 0.016$\,\%$ \\
\midrule
\multicolumn{3}{l}{\textit{Inference cost}} \\
\quad Inference time (ms)         &  72.61 &  53.41 \\
\quad Inference peak memory (MiB) & 5{,}869 & 2{,}918 \\
\midrule
\multicolumn{3}{l}{\textit{Adaptation cost (per training step)}} \\
\quad Step time (ms)              & 1{,}616 &    418 \\
\quad Step peak memory (MiB)      & 7{,}335 & 8{,}392 \\
\bottomrule
\end{tabular}
\vspace{-10pt}

\end{table}

\section{Compute Resources and Efficiency Analysis}
\label{appendix:compute}

In Table~\ref{tab:efficiency}, we report the compute and memory profile of our adaptation framework 
on a single NVIDIA H200 GPU, using BDD100K (998 validation images, 
19 classes) as a representative target.

\subsection{Trainable Parameters}
\label{appendix:trainable-params}

Our adaptation introduces only two trainable modules per backbone: 
a vision LoRA on the last four CLIP-ViT blocks (Q and V projections, 
rank 4) and a rank-4 residual text adapter shared across all classes 
and templates. The trainable footprint is \textbf{71{,}681 parameters} 
on both CAT-Seg-L and SAN-L (65{,}536 vision LoRA $+$ 6{,}145 text 
adapter), corresponding to roughly $0.017\%$ of the full model. The 
footprint is essentially identical because both adapter modules 
attach to shared CLIP components, independent of the surrounding 
OVSS architecture.

\subsection{Adaptation Cost}
\label{appendix:adaptation-cost}

We measure the additional cost incurred by our adaptation as the 
gap between a base inference forward and a single training step, 
where each step performs one update on a support image including 
the $K = 14$ candidate forwards used for winner/loser selection. The 
adaptation overhead is $+1{,}466$\,MiB and $+1{,}543$\,ms on 
CAT-Seg-L, and $+5{,}474$\,MiB and $+365$\,ms on SAN-L. CAT-Seg-L incurs a smaller memory 
overhead but a longer per-step time because its IoU-based 
winner/loser selection requires one full sliding-window forward per 
template, 
whereas SAN-L scores all $K = 14$ candidates with native single-pass 
forwards.

\section{Additional Qualitative Results}
\label{appendix:additional-qual}

\begin{figure}[t]
\centering
\includegraphics[width=\linewidth]{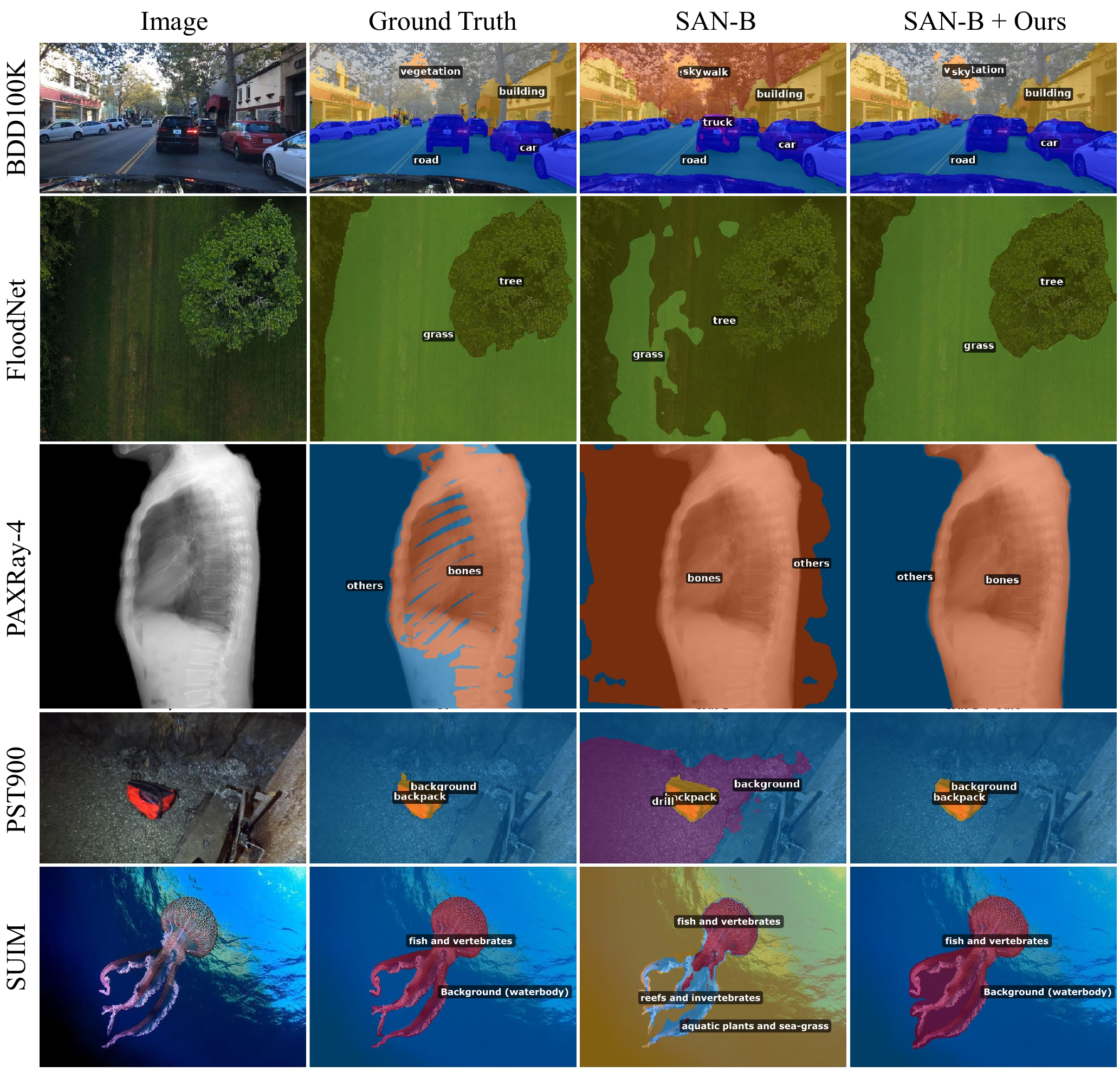}
\caption{Qualitative results on SAN-B. Each row shows one sample 
from one of the five MESS domain groups (top to bottom: General, 
Earth Monitoring, Medical Sciences, Engineering, Agriculture \& 
Biology). Columns show the input image, ground-truth 
segmentation, zero-shot baseline 
prediction, and our adapted prediction.}
\label{fig:qual-san-b}
\end{figure}

\begin{figure}[t]
\centering
\includegraphics[width=\linewidth]{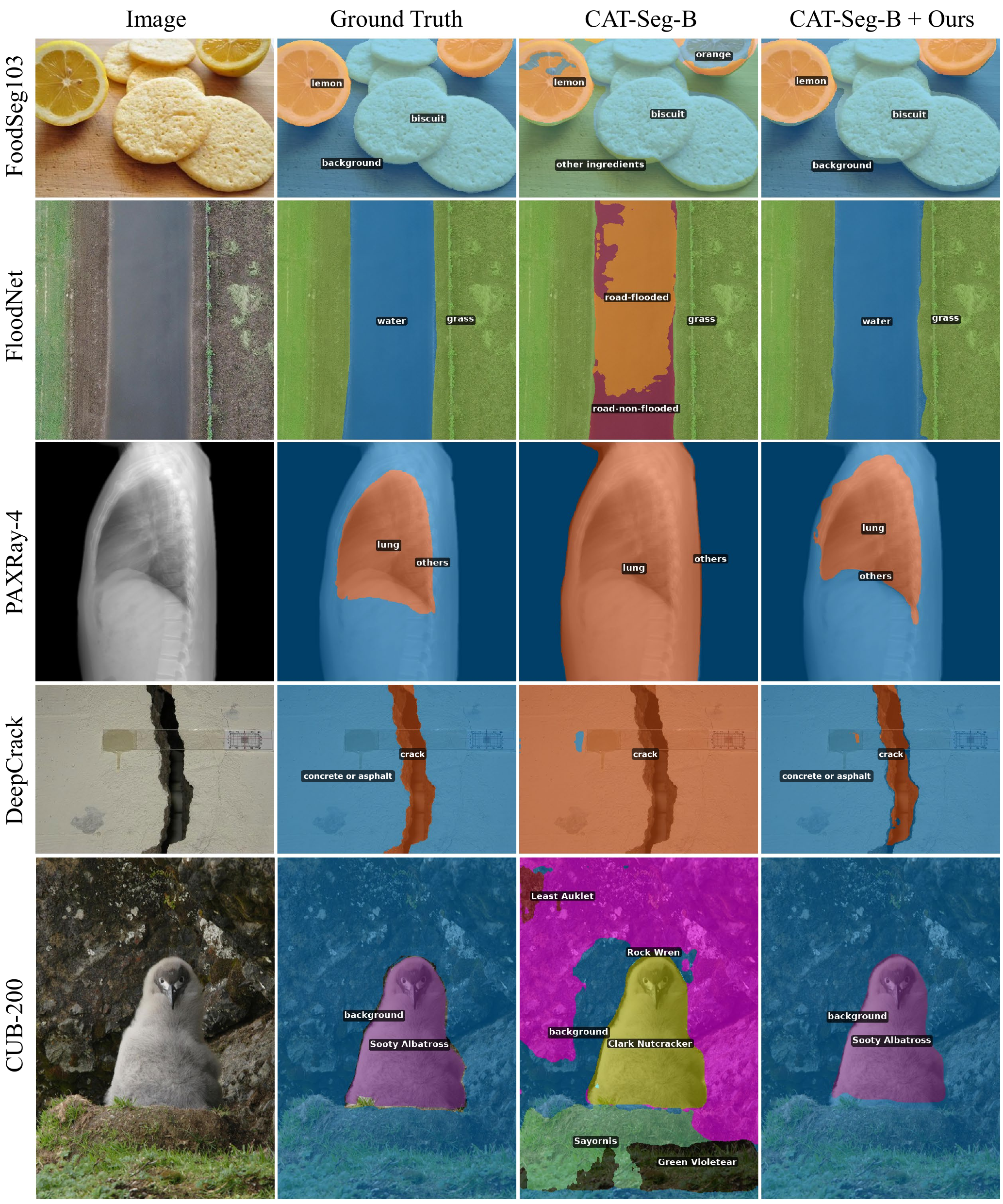}
\caption{Qualitative results on CAT-Seg-B. Each row shows one sample 
from one of the five MESS domain groups (top to bottom: General, 
Earth Monitoring, Medical Sciences, Engineering, Agriculture \& 
Biology). Columns show the input image, ground-truth 
segmentation, zero-shot baseline 
prediction, and our adapted prediction.}
\label{fig:qual-cat-seg-b}
\end{figure}

\begin{figure}[t]
\centering
\includegraphics[width=\linewidth]{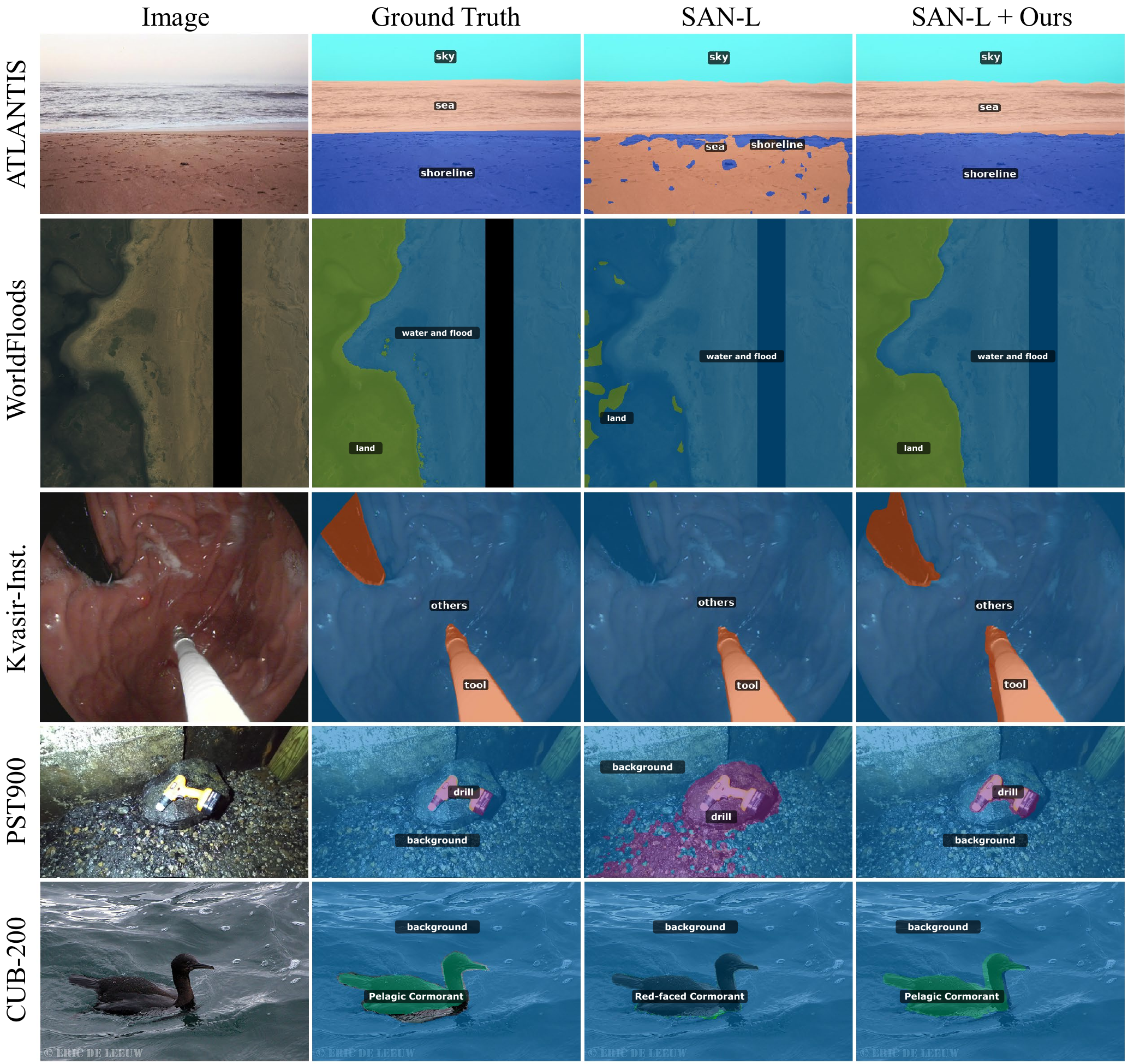}
\caption{Qualitative results on SAN-L. Each row shows one sample 
from one of the five MESS domain groups (top to bottom: General, 
Earth Monitoring, Medical Sciences, Engineering, Agriculture \& 
Biology). Columns show the input image, ground-truth 
segmentation, zero-shot baseline 
prediction, and our adapted prediction.}
\label{fig:qual-san-l}
\end{figure}

\begin{figure}[t]
\centering
\includegraphics[width=\linewidth]{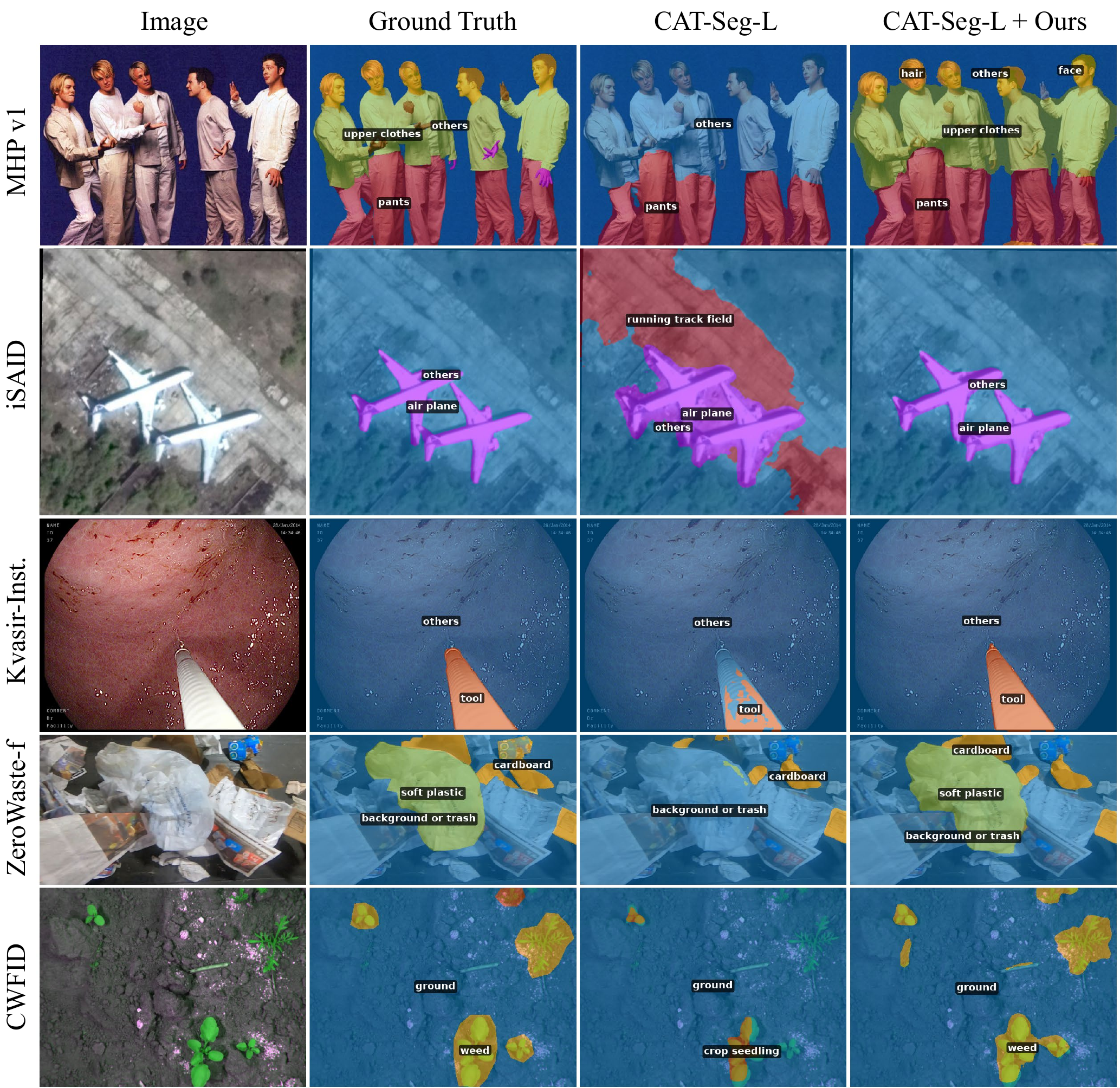}
\caption{Qualitative results on CAT-Seg-L. Each row shows one sample 
from one of the five MESS domain groups (top to bottom: General, 
Earth Monitoring, Medical Sciences, Engineering, Agriculture \& 
Biology). Columns show the input image, ground-truth 
segmentation, zero-shot baseline 
prediction, and our adapted prediction.}
\label{fig:qual-cat-seg-l}
\end{figure}

We present additional qualitative results.
Figures~\ref{fig:qual-san-b}--\ref{fig:qual-cat-seg-l} show 
per-backbone adaptation results across the five MESS domain 
groups: for each backbone (SAN-B, CAT-Seg-B, SAN-L, CAT-Seg-L), 
we visualize one sample per domain group, comparing the input 
image, the zero-shot baseline prediction, our adapted prediction, 
and the ground-truth segmentation. 

\begin{figure}[t]
\centering
\includegraphics[width=\linewidth]{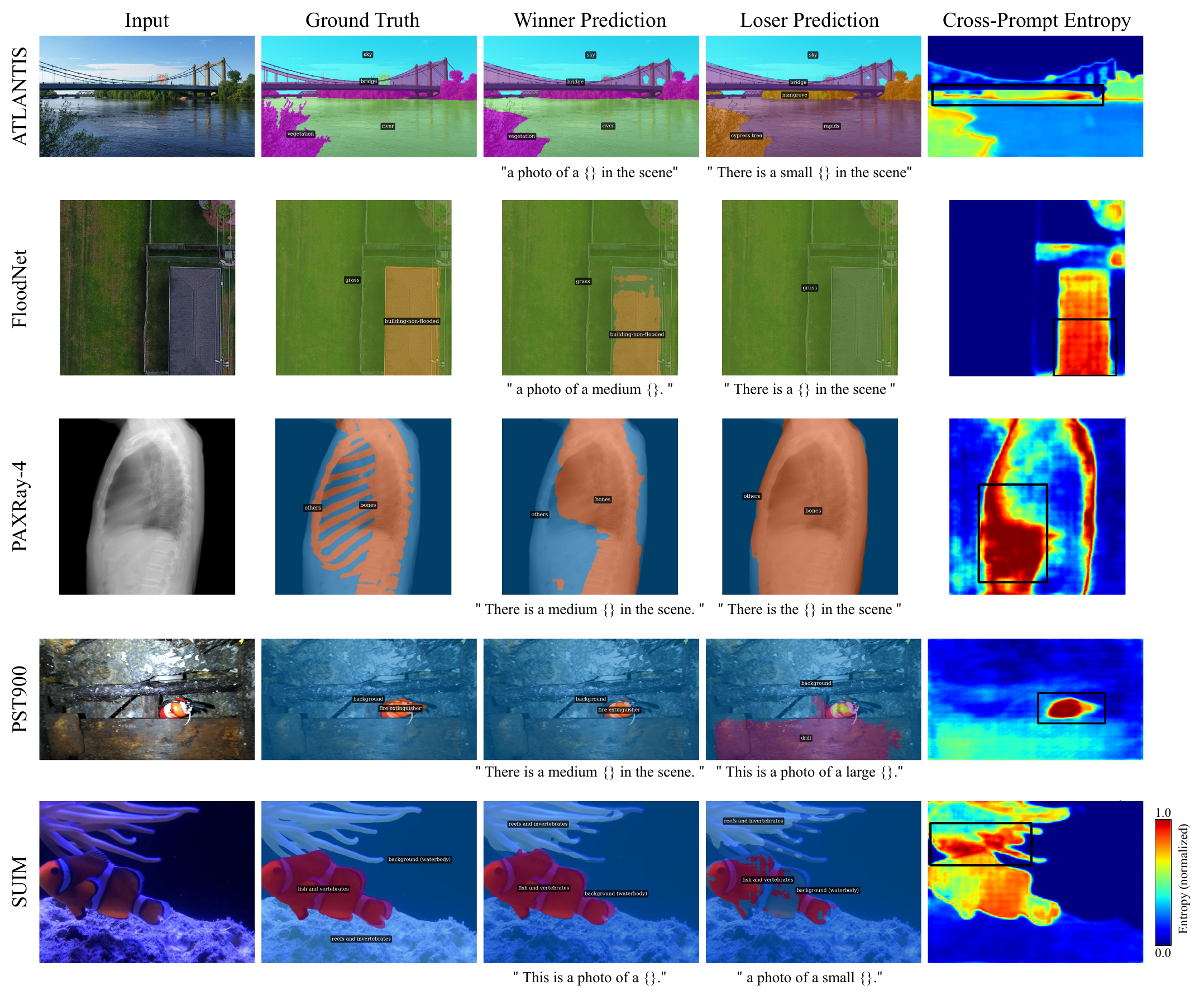}
\caption{Visualization of preference query mining. For each example 
(one per row), we show the input image, ground truth, winner 
prediction, loser prediction, and cross-template entropy 
($K = 14$ ViLD templates). The selected query region $\mathcal{R}$ 
is overlaid on the entropy map as a bounding box.}
\label{fig:disagreement-viz}
\end{figure}

Figure~\ref{fig:disagreement-viz} visualizes the preference query 
mining process described in Section~\ref{sec:query}. For 
each example, we show the input image, the cross-prompt 
uncertainty map (per-pixel ensemble entropy across the $K = 14$ 
templates), and the selected query region $\mathcal{R}$ overlaid 
on the prediction as a bounding box. The high-entropy regions 
concentrate on object boundaries and parts of the image where 
templates disagree most strongly, illustrating how prompt 
disagreement provides a localized, informative signal for 
preference query mining.
\clearpage

\begin{table}[h]
\centering
\caption{The $K = 14$ prompt templates from the ViLD prompt pool used 
throughout this work.}
\vspace{5pt}
\label{tab:vild-templates}
\begin{tabular}{cl}
\toprule
\# & Template \\
\midrule
1 & a photo of a \{\}. \\
2 & This is a photo of a \{\}. \\
3 & There is a \{\} in the scene. \\
4 & There is the \{\} in the scene. \\
5 & a photo of a \{\} in the scene. \\
6 & a photo of a small \{\}. \\
7 & a photo of a medium \{\}. \\
8 & a photo of a large \{\}. \\
9 & This is a photo of a small \{\}. \\
10 & This is a photo of a medium \{\}. \\
11 & This is a photo of a large \{\}. \\
12 & There is a small \{\} in the scene. \\
13 & There is a medium \{\} in the scene. \\
14 & There is a large \{\} in the scene. \\
\bottomrule
\end{tabular}
\end{table}

\begin{table}[h]
\centering
\caption{Datasets in the MESS benchmark used in this work, grouped by 
domain.}
\label{tab:dataset-detail}
\small
\resizebox{\linewidth}{!}{%
\begin{tabular}{llcl}
\toprule
Dataset & License   & \# classes & Classes \\
\midrule
\multicolumn{4}{l}{\textit{General Scenes}} \\
\midrule
BDD100K     ~\cite{DatasetBDD100K}            & custom                  & 19  & [road; sidewalk; building; wall; fence; pole; traffic light; \ldots] \\
MHP v1      ~\cite{DatasetMHPv1}              & custom                  & 19  & [others; hat; hair; sunglasses; upper clothes; skirt; pants; \ldots] \\
FoodSeg103  ~\cite{DatasetFoodSeg103}         & Apache 2.0              & 104 & [background; candy; egg tart; french fries; chocolate; biscuit; \ldots] \\
ATLANTIS    ~\cite{DatasetATLANTIS}           & Flickr (images)  & 56  & [bicycle; boat; breakwater; bridge; building; bus; canal; \ldots] \\
\midrule
\multicolumn{4}{l}{\textit{Earth Monitoring}} \\
\midrule
iSAID       ~\cite{DatasetiSAID}              & Google Earth (images)   & 16  & [others; boat; storage tank; baseball diamond; tennis court; \ldots] \\
WorldFloods ~\cite{DatasetWorldFloods}        & CC NC 4.0               & 3   & [land; water and flood; cloud] \\
FloodNet    ~\cite{DatasetFloodNet}           & custom                  & 10  & [building-flooded; building-non-flooded; road-flooded; water; \ldots] \\
UAVid       ~\cite{DatasetUAVid}              & CC BY-NC-SA 4.0         & 8   & [others; building; road; tree; grass; moving car; parked car; humans] \\
\midrule
\multicolumn{4}{l}{\textit{Medical Sciences}} \\
\midrule
Kvasir-Instrument ~\cite{DatasetKvasirInstrument}   & custom                  & 2   & [others; tool] \\
CHASE\_DB1  ~\cite{DatasetCHASEDB1}           & CC BY 4.0               & 2   & [others; blood vessels] \\
PAXRay-4    ~\cite{DatasetPAXRay4}            & custom                  & 4$\times$2 & [others, lungs], [others, bones], [others, mediastinum], [others, diaphragm] \\

\midrule
\multicolumn{4}{l}{\textit{Engineering}} \\
\midrule
Corrosion CS~\cite{DatasetCorrosionCS}        & CC0                     & 4   & [others; steel with fair, poor, severe corrosion] \\
DeepCrack   ~\cite{DatasetDeepCrack}          & custom                  & 2   & [concrete or asphalt; crack] \\
PST900      ~\cite{DatasetPST900}             & GPL-3.0                 & 5   & [background; fire extinguisher; backpack; drill; human] \\
ZeroWaste-f ~\cite{DatasetZeroWastef}         & CC-BY-NC 4.0            & 5   & [background or trash; rigid plastic; cardboard; metal; soft plastic] \\

\midrule
\multicolumn{4}{l}{\textit{Agriculture \& Biology}} \\
\midrule
SUIM          ~\cite{DatasetSUIM}               & MIT                     & 8   & [human diver; reefs and invertebrates; fish and vertebrates; \ldots] \\
CUB-200     ~\cite{DatasetCUB200}             & custom                  & 201 & [background; Laysan Albatross; Sooty Albatross; Crested Auklet; \ldots] \\
CWFID       ~\cite{DatasetCWFID}              & custom                  & 3   & [ground; crop seedling; weed] \\

\bottomrule
\end{tabular}
}
\end{table}
\setcounter{topnumber}{4}
\setcounter{totalnumber}{4}
\clearpage
\begin{table}[h]
\centering
\caption{Per-dataset results on the MESS benchmark across four OVSS 
backbones (mIoU, \%). Within each backbone, the better of the 
zero-shot baseline and our method (\textit{+ Ours}) is highlighted 
in bold. (\textit{+ Dense.}) indicates the supervised reference.}
\vspace{2pt}
\label{tab:per-dataset-main}
\small
\setlength{\tabcolsep}{3pt}
\resizebox{\linewidth}{!}{%
\begin{tabular}{ll cccc cccc ccc cccc ccc| c}
\toprule
& & \multicolumn{4}{c}{General} & \multicolumn{4}{c}{Earth Monit.} & \multicolumn{3}{c}{Medical} & \multicolumn{4}{c}{Engineering} & \multicolumn{3}{c}{Agri.\ \& Bio.} & \\
\cmidrule(lr){3-6} \cmidrule(lr){7-10} \cmidrule(lr){11-13} \cmidrule(lr){14-17} \cmidrule(lr){18-20}
Backbone & Method
& \rotatebox{90}{BDD100K} & \rotatebox{90}{MHP v1} & \rotatebox{90}{FoodSeg103} & \rotatebox{90}{ATLANTIS}
& \rotatebox{90}{iSAID} & \rotatebox{90}{WorldFloods} & \rotatebox{90}{FloodNet} & \rotatebox{90}{UAVid}
& \rotatebox{90}{Kvasir-Inst.} & \rotatebox{90}{CHASE\_DB1} & \rotatebox{90}{PAXRay-4}
& \rotatebox{90}{Corrosion CS} & \rotatebox{90}{DeepCrack} & \rotatebox{90}{PST900} & \rotatebox{90}{ZeroWaste-f}
& \rotatebox{90}{SUIM} & \rotatebox{90}{CUB-200} & \rotatebox{90}{CWFID}
& \textit{Mean} \\
\midrule
\multirow{3}{*}{SAN-B}
& base    & 35.36 & 9.39 & \textbf{8.40} & 33.25 & 4.18 & 30.15 & \textbf{33.95} & 39.31 & 62.38 & 18.47 & 19.69 & 4.53 & \textbf{49.27} & 40.69 & \textbf{18.25} & 36.54 & 5.79 & 5.63 & \textit{25.29} \\
& + Ours  & \textbf{37.78} & \textbf{10.69} & 6.59 & \textbf{33.31} & \textbf{4.19} & \textbf{32.59} & 33.73 & \textbf{40.25} & \textbf{73.30} & \textbf{46.43} & \textbf{39.18} & \textbf{21.01} & 48.23 & \textbf{46.49} & 18.16 & \textbf{40.64} & \textbf{6.77} & \textbf{43.68} & \textit{\textbf{32.39}} \\
& + Dense.  & 40.04 & 12.40 & 12.77 & 33.75 & 6.70 & 33.30 & 34.70 & 39.66 & 47.71 & 46.68 & 38.71 & 21.01 & 61.59 & 42.20 & 17.76 & 49.58 & 6.82 & 38.05 & \textit{32.41} \\
\midrule
\multirow{3}{*}{CAT-Seg-B}
& base    & 47.03 & 23.89 & 26.67 & \textbf{40.43} & 19.34 & \textbf{38.52} & 37.16 & 43.04 & 48.20 & \textbf{23.99} & 41.26 & 12.46 & 32.71 & 57.13 & 17.51 & 44.82 & \textbf{10.41} & 31.61 & \textit{33.12} \\
& + Ours  & \textbf{48.12} & \textbf{29.00} & \textbf{27.78} & 39.35 & \textbf{24.70} & 36.60 & \textbf{37.91} & \textbf{44.29} & \textbf{84.05} & 23.16 & \textbf{50.77} & \textbf{22.33} & \textbf{61.49} & \textbf{73.79} & \textbf{17.65} & \textbf{47.08} & 9.43 & \textbf{36.60} & \textit{\textbf{39.67}} \\
& + Dense.  & 48.93 & 34.69 & 29.67 & 41.48 & 24.14 & 39.57 & 40.18 & 49.00 & 77.39 & 46.65 & 56.21 & 26.67 & 56.11 & 75.31 & 25.94 & 63.58 & 14.98 & 46.26 & \textit{44.26} \\
\midrule
\multirow{3}{*}{SAN-L}
& base    & 42.65 & 9.16 & 14.50 & 38.61 & 10.31 & \textbf{48.89} & 37.42 & 41.42 & 62.25 & 4.43 & 29.33 & 8.17 & 19.65 & 53.73 & 15.03 & \textbf{48.03} & 8.27 & 1.43 & \textit{27.40} \\
& + Ours  & \textbf{44.68} & \textbf{11.17} & \textbf{15.23} & \textbf{38.63} & \textbf{19.38} & 47.95 & \textbf{38.56} & \textbf{42.45} & \textbf{67.50} & \textbf{5.97} & \textbf{42.40} & \textbf{20.88} & \textbf{47.84} & \textbf{59.38} & \textbf{16.41} & 47.07 & \textbf{8.84} & \textbf{37.54} & \textit{\textbf{33.99}} \\
& + Dense.  & 46.75 & 14.91 & 23.86 & 39.19 & 22.77 & 37.75 & 38.51 & 43.91 & 85.10 & 15.47 & 49.59 & 21.04 & 47.84 & 63.29 & 19.96 & 49.13 & 9.78 & 48.68 & \textit{37.64} \\
\midrule
\multirow{3}{*}{CAT-Seg-L}
& base    & 48.23 & 30.77 & 32.92 & \textbf{45.51} & 19.67 & 39.94 & 41.05 & 41.90 & 65.49 & 3.32 & 19.75 & 7.47 & 25.27 & 78.73 & 25.42 & 49.75 & 21.89 & 37.58 & \textit{35.26} \\
& + Ours  & \textbf{50.58} & \textbf{36.41} & \textbf{37.41} & 45.50 & \textbf{36.17} & \textbf{41.13} & \textbf{41.07} & \textbf{42.74} & \textbf{87.41} & \textbf{20.74} & \textbf{47.34} & \textbf{22.90} & \textbf{74.39} & \textbf{80.96} & \textbf{32.47} & \textbf{58.49} & \textbf{22.22} & \textbf{47.87} & \textit{\textbf{45.88}} \\
& + Dense.  & 51.59 & 39.03 & 38.57 & 46.97 & 31.51 & 40.63 & 43.63 & 51.49 & 85.13 & 21.45 & 59.85 & 22.74 & 62.14 & 78.68 & 33.17 & 59.01 & 26.93 & 47.49 & \textit{46.67} \\
\bottomrule
\end{tabular}%
}
\vspace{-18pt}
\end{table}
\begin{table}[h]
\centering
\caption{Per-dataset loss ablation. The best performance is highlighted in bold.}
\vspace{2pt}
\label{tab:per-dataset-loss}
\small
\setlength{\tabcolsep}{3pt}
\resizebox{\linewidth}{!}{%
\begin{tabular}{l cccc cccc ccc cccc ccc| c}
\toprule
& \multicolumn{4}{c}{General} & \multicolumn{4}{c}{Earth Monit.} & \multicolumn{3}{c}{Medical} & \multicolumn{4}{c}{Engineering} & \multicolumn{3}{c}{Agri.\ \& Bio.} & \\
\cmidrule(lr){2-5} \cmidrule(lr){6-9} \cmidrule(lr){10-12} \cmidrule(lr){13-16} \cmidrule(lr){17-19}
Method
& \rotatebox{90}{BDD100K} & \rotatebox{90}{MHP v1} & \rotatebox{90}{FoodSeg103} & \rotatebox{90}{ATLANTIS}
& \rotatebox{90}{iSAID} & \rotatebox{90}{WorldFloods} & \rotatebox{90}{FloodNet} & \rotatebox{90}{UAVid}
& \rotatebox{90}{Kvasir-Inst.} & \rotatebox{90}{CHASE\_DB1} & \rotatebox{90}{PAXRay-4}
& \rotatebox{90}{Corrosion CS} & \rotatebox{90}{DeepCrack} & \rotatebox{90}{PST900} & \rotatebox{90}{ZeroWaste-f}
& \rotatebox{90}{SUIM} & \rotatebox{90}{CUB-200} & \rotatebox{90}{CWFID}
& \textit{Mean} \\
\midrule
Baseline                            & 48.23 & 30.77 & 32.92 & 45.51 & 19.67 & 39.94 & 41.05 & 41.90 & 65.49 & 3.32 & 19.75 & 7.47 & 25.27 & 78.73 & 25.42 & 49.75 & 21.89 & 37.58 & \textit{35.26} \\
w/o $\mathcal{L}_{\text{RLPO}}$     & 50.08 & 35.41 & 36.65 & 45.05 & 31.67 & 40.83 & 40.01 & 41.56 & 70.68 & 3.32 & 34.76 & 20.88 & 61.25 & 80.62 & 16.57 & 55.04 & 21.54 & 43.32 & \textit{40.51} \\
w/o $\mathcal{L}_{\text{cons}}$     & 50.36 & 34.48 & 36.79 & \textbf{45.56} & 25.45 & \textbf{41.29} & \textbf{42.07} & \textbf{43.64} & 84.86 & 20.52 & 41.40 & \textbf{23.78} & 59.33 & 80.22 & 28.18 & 54.90 & 21.15 & \textbf{48.17} & \textit{43.45} \\
Ours                                & \textbf{50.58} & \textbf{36.41} & \textbf{37.41} & 45.50 & \textbf{36.17} & 41.13 & 41.07 & 42.74 & \textbf{87.41} & \textbf{20.74} & \textbf{47.34} & 22.90 & \textbf{74.39} & \textbf{80.96} & \textbf{32.47} & \textbf{58.49} & \textbf{22.22} & 47.87 & \textit{\textbf{45.88}} \\
\bottomrule
\end{tabular}%
}
\vspace{-18pt}
\end{table}
\begin{table}[h]
\centering
\caption{Per-dataset candidate generation comparison. With matched candidate size ($K = 14$), we contrast prompt disagreement (Ours) against MC Dropout and test-time 
augmentation (TTA). The best performance is highlighted in bold.}
\vspace{2pt}
\label{tab:per-dataset-candidate}
\small
\setlength{\tabcolsep}{3pt}
\resizebox{\linewidth}{!}{%
\begin{tabular}{l cccc cccc ccc cccc ccc| c}
\toprule
& \multicolumn{4}{c}{General} & \multicolumn{4}{c}{Earth Monit.} & \multicolumn{3}{c}{Medical} & \multicolumn{4}{c}{Engineering} & \multicolumn{3}{c}{Agri.\ \& Bio.} & \\
\cmidrule(lr){2-5} \cmidrule(lr){6-9} \cmidrule(lr){10-12} \cmidrule(lr){13-16} \cmidrule(lr){17-19}
Method
& \rotatebox{90}{BDD100K} & \rotatebox{90}{MHP v1} & \rotatebox{90}{FoodSeg103} & \rotatebox{90}{ATLANTIS}
& \rotatebox{90}{iSAID} & \rotatebox{90}{WorldFloods} & \rotatebox{90}{FloodNet} & \rotatebox{90}{UAVid}
& \rotatebox{90}{Kvasir-Inst.} & \rotatebox{90}{CHASE\_DB1} & \rotatebox{90}{PAXRay-4}
& \rotatebox{90}{Corrosion CS} & \rotatebox{90}{DeepCrack} & \rotatebox{90}{PST900} & \rotatebox{90}{ZeroWaste-f}
& \rotatebox{90}{SUIM} & \rotatebox{90}{CUB-200} & \rotatebox{90}{CWFID}
& \textit{Mean} \\
\midrule
Baseline    & 48.23 & 30.77 & 32.92 & 45.51 & 19.67 & 39.94 & 41.05 & 41.90 & 65.49 & 3.32 & 19.75 & 7.47 & 25.27 & 78.73 & 25.42 & 49.75 & 21.89 & 37.58 & \textit{35.26} \\
MC Dropout  & 47.59 & 17.67 & 35.41 & 45.23 & 18.63 & 29.23 & 38.28 & 42.27 & 81.48 & 15.01 & 31.43 & \textbf{23.22} & 70.32 & 77.16 & 19.93 & 45.85 & 21.33 & 43.52 & \textit{39.09} \\
TTA         & 47.59 & 34.41 & 35.78 & \textbf{46.01} & 27.12 & 40.02 & \textbf{41.11} & 42.66 & \textbf{88.12} & \textbf{21.06} & \textbf{50.28} & 22.30 & 72.84 & 74.21 & \textbf{33.36} & 57.35 & \textbf{22.56} & 47.17 & \textit{44.66} \\
Ours        & \textbf{50.58} & \textbf{36.41} & \textbf{37.41} & 45.50 & \textbf{36.17} & \textbf{41.13} & 41.07 & \textbf{42.74} & 87.41 & 20.74 & 47.34 & 22.90 & \textbf{74.39} & \textbf{80.96} & 32.47 & \textbf{58.49} & 22.22 & \textbf{47.87} & \textit{\textbf{45.88}} \\
\bottomrule
\end{tabular}%
}
\vspace{-18pt}

\end{table}
\begin{table}[h]
\centering
\caption{Per-dataset effect of training set size. The best performance is highlighted in bold.}
\vspace{2pt}
\label{tab:per-dataset-shot}
\small
\setlength{\tabcolsep}{3pt}
\resizebox{\linewidth}{!}{%
\begin{tabular}{c cccc cccc ccc cccc ccc| c}
\toprule
& \multicolumn{4}{c}{General} & \multicolumn{4}{c}{Earth Monit.} & \multicolumn{3}{c}{Medical} & \multicolumn{4}{c}{Engineering} & \multicolumn{3}{c}{Agri.\ \& Bio.} & \\
\cmidrule(lr){2-5} \cmidrule(lr){6-9} \cmidrule(lr){10-12} \cmidrule(lr){13-16} \cmidrule(lr){17-19}
\# images
& \rotatebox{90}{BDD100K} & \rotatebox{90}{MHP v1} & \rotatebox{90}{FoodSeg103} & \rotatebox{90}{ATLANTIS}
& \rotatebox{90}{iSAID} & \rotatebox{90}{WorldFloods} & \rotatebox{90}{FloodNet} & \rotatebox{90}{UAVid}
& \rotatebox{90}{Kvasir-Inst.} & \rotatebox{90}{CHASE\_DB1} & \rotatebox{90}{PAXRay-4}
& \rotatebox{90}{Corrosion CS} & \rotatebox{90}{DeepCrack} & \rotatebox{90}{PST900} & \rotatebox{90}{ZeroWaste-f}
& \rotatebox{90}{SUIM} & \rotatebox{90}{CUB-200} & \rotatebox{90}{CWFID}
& \textit{Mean} \\
\midrule
0   & 48.23 & 30.77 & 32.92 & \textbf{45.51} & 19.67 & 39.94 & 41.05 & 41.90 & 65.49 &  3.32 & 19.75 &  7.47 & 25.27 & 78.73 & 25.42 & 49.75 & 21.89 & 37.58 & \textit{35.26} \\
4   & 49.59 & 31.71 & 33.30 & 43.69 & 29.30 & 39.96 & 41.04 & 42.57 & 68.67 & 20.24 & 34.17 & 15.21 & 26.02 & 79.57 & 24.33 & 52.39 & 18.64 & 38.24 & \textit{38.26} \\
8   & 50.24 & 32.36 & 34.10 & 45.40 & 26.51 & 39.78 & 41.14 & 42.10 & 69.34 & 20.74 & 38.19 & 23.04 & 29.24 & 79.86 & 24.63 & 52.54 & 21.08 & 40.75 & \textit{39.50} \\
16  & 50.73 & 31.89 & 35.67 & 45.13 & 34.12 & 40.47 & 41.22 & 42.20 & 74.79 & 20.74 & 43.38 & 22.89 & 53.77 & 80.58 & 23.53 & 54.73 & 19.25 & 46.51 & \textit{42.31} \\
32  & 50.78 & 34.43 & 37.20 & 45.43 & 30.62 & 40.27 & \textbf{41.25} & 42.63 & 80.26 & 20.74 & 45.68 & 22.24 & 72.25 & 80.31 & 27.46 & 53.58 & 22.64 & 47.85 & \textit{44.20} \\
64  & 50.58 & 36.41 & \textbf{37.41} & 45.50 & \textbf{36.17} & 41.13 & 41.07 & \textbf{42.74} & \textbf{87.41} & 20.74 & 47.34 & 22.90 & 74.39 & \textbf{80.96} & 32.47 & \textbf{58.49} & 22.22 & \textbf{47.87} & \textit{\textbf{45.88}} \\
128 & \textbf{51.24} & \textbf{36.71} & 36.83 & 45.47 & 33.85 & \textbf{41.34} & 40.76 & 41.44 & 85.21 & 20.74 & \textbf{48.30} & \textbf{25.47} & \textbf{75.99} & 80.49 & \textbf{34.59} & 54.53 & \textbf{23.41} & 47.85 & \textit{45.79} \\
\bottomrule
\end{tabular}%
}
\vspace{-10pt}
\end{table}
\clearpage

\end{document}